# Dynamically Allocating Evaluation Effort for Model Ranking

**Vilém Zouhar** ETH Zurich

**Julia Kreutzer** Cohere Labs

**Alon Lavie** Carnegie Mellon University

**Tom Kocmi** Cohere

**Matt Post** Microsoft

**Ondřej Bojar** Charles University

**Mrinmaya Sachan** ETH Zurich

## Abstract

While human evaluation is the gold standard in many NLP tasks, it suffers from prohibitive costs and poor scalability. When identifying top-performing models, typical evaluation protocols waste effort by exhaustively evaluating all models on the entire benchmark, a safe but inefficient approach. In this work, we formalize multi-model human evaluation as a best-arm identification problem in a multi-armed bandit setup with correlated arms, where pulling an arm corresponds to human-evaluating a model. By sampling adaptively based on the intermediate model rankings obtained on the samples so far, we can focus the annotation budget on the most competitive models. We prove the optimality of the proposed algorithms and show that it improves discrimination between top-performing models. This makes evaluations faster, cheaper and more aligned with large-scale competition evaluation goals.[1]

## 1 Introduction

Human evaluation guides research and deployment decisions, but quickly grows unsustainable due to the massive budget required to obtain exhaustive ranking of models (Graham et al., 2013b; Kocmi et al., 2025). In the absence of human supervision, researchers rely on automatic metrics which are often only modestly correlated with quality, are susceptible to reward-hacking, and are less accurate for state-of-the-art models (Lavie et al., 2025). To this end, (Graham et al., 2013a; Freitag et al., 2021; Kocmi et al., 2024b) make the annotation process itself more robust, and (Zouhar et al., 2025; Proietti et al., 2025) improve the overall evaluation efficiency by prioritizing informative evaluation items. Unfortunately, these efforts operate within a static framework where all models are evaluated on all items in the testset, instead of strategically allocating evaluation effort where it is most needed (e.g. top performing and close matching models).

We frame allocation of evaluation effort as a search problem and introduce a dynamic evaluation policy that treats evaluation as an expensive resource allocated via a multi-armed bandit framework. By dynamically selecting which models to next evaluate, we automatically concentrate evaluation effort on top-performing models while allowing for exploration of all models, which is ultimately more useful in determining the best model. As illustrated in Figure 1, this contrasts with standard uniform allocation: by focusing on top models, we direct the annotation budget to where it matters more. In this work we:

- formalize allocation of evaluation effort as a best-arm identification (Section 3), where pulling an arm(=model) corresponds to evaluating it to more items, and show newly NP-hardness even with an oracle,
- evaluate existing and propose new simple allocation algorithms (Section 4), such as selecting

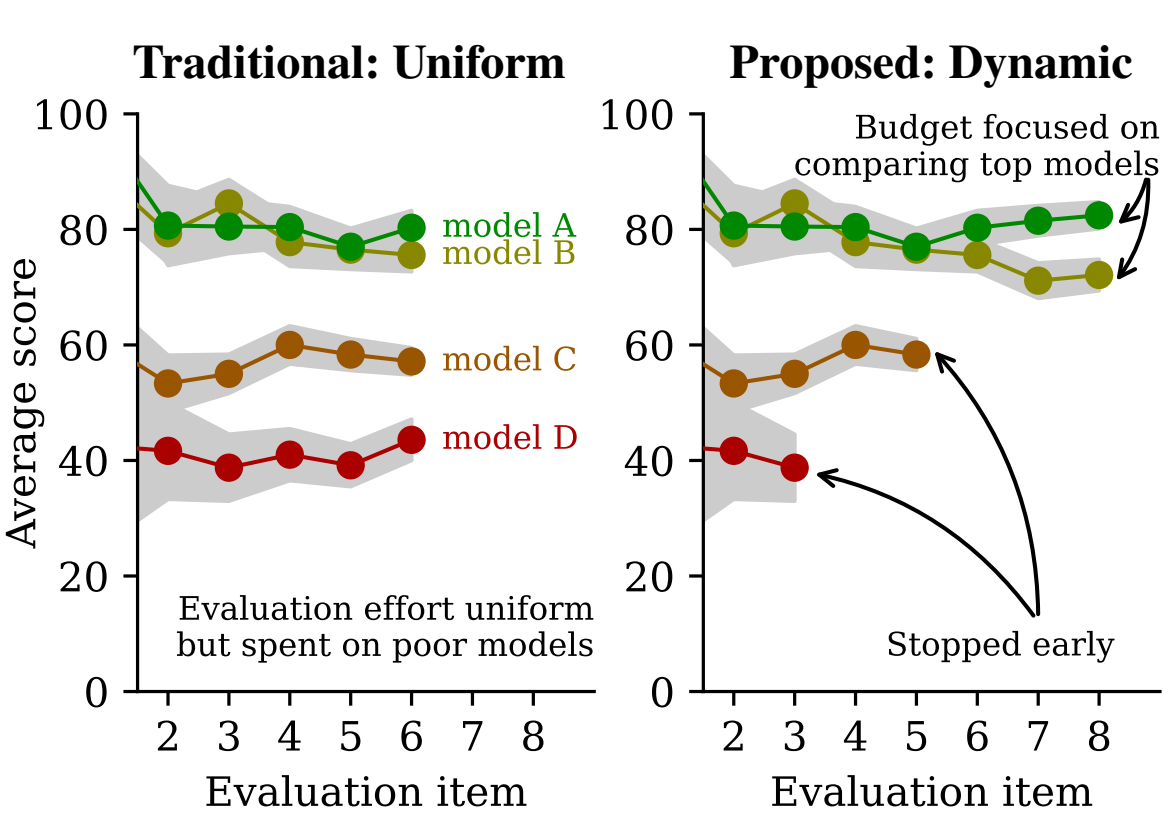

Figure 1: Classical evaluation vs our proposed approach: we evaluate the same number of items but in dynamic evaluation we automatically focus on top-performing models. This informs a better choice of close-matching state-of-the-art models.

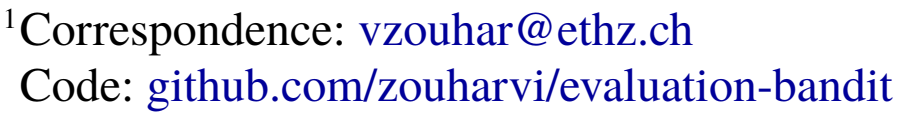

[1]Correspondence: vzouhar@ethz.ch
Code: github.com/zouharvi/evaluation-bandit

to evaluate model with probability inverse to its currently predicted rank, which we newly prove to be optimal under certain assumptions,
- simulate dynamic evaluation for large-scale translation benchmarking (Section 5), demonstrating that dynamic allocation yields superior rankings among top models.
- incorporate statistical models to compensate for biased evaluation item ordering (Section 3.4),

Our approach unlocks faster and cheaper large-scale evaluations with many competing models, and is applicable to benchmarking shared-tasks, model selection during development, and scenarios where the cost of even automatic evaluation becomes the bottleneck. We confirm our theoretical results by a case study on large-scale translation benchmarking (WMT; Kocmi et al., 2025).

## 2 Related Work

Human evaluation is the gold standard used to compare and rank models, but is expensive, which is a key bottleneck of modern benchmarking. Methods of efficient evaluation apply also to evaluations done automatically, where the cost does not correspond to annotators' time, but to compute or API costs, such as when evaluating with LLM-as-a-judge.

**Matchmaking and model evaluation.** A common approach for human evaluation where objective assessments are not possible is to collect pairwise comparisons. Mathematical models, such as ELO or TrueSkill, both predict which models should be matched(=evaluated) together and then global model rankings are inferred from noisy pairwise annotations (Elo, 1978; Herbrich et al., 2006; Minka et al., 2018). Recent work adapts this paradigm to language models: Chatbot Arena aggregates open-ended pairwise judgments via ELO (Zheng et al., 2023), while K-Sort Arena employs bandit-style exploration to improve evaluation efficiency (Li et al., 2025b).

These settings assume that evaluations across models are independent due to free-form inputs rather than fixed items from a dataset, which is our focus. Furthermore, standard approaches aim to best estimate the full model ranking (Sakaguchi and Van Durme, 2018; Balkır et al., 2026). However, our objective is not to rank every model, but to rapidly discard weak candidates and concentrate the budget on resolving uncertainty among top models.

**Evaluation item selection.** A separate line of work, distinct from allocating which *model* is to be evaluated, optimizes which *items* should be evaluated. Subset selection methods approximate full-benchmark rankings using a small budget of difficult or otherwise informative items (Polo et al., 2024; Ni et al., 2024; Saldías Fuentes et al., 2022; Zouhar et al., 2025; Proietti et al., 2025). These strategies typically assume a uniform allocation of effort across models and optimize *item* choice. However, we show that smart evaluation item selection approaches are compatible with selecting which models to evaluate.

See Appendix E for an extended related work.

## 3 Which Models and Items to Evaluate?

We begin by formalizing the challenge of efficient evaluation. We then describe a set of simplifying assumptions, focusing on selecting which *model* to evaluate next, separating it from the problem of selecting evaluation *items*, which is part of several other works.[2]

### 3.1 Problem statement

Let $\mathcal{M}$ denote the set of models to evaluate and compare, $\mathcal{X}$ denote a set of evaluation items, and $B \in \mathbb{R}^+$ denote the given evaluation budget. Given the output of model $m$ on item $x$, let $r_{x,m} \in [0,1]$ denote the evaluation score and $\text{cost}(x) \in \mathbb{R}^+$ denote the cost of this evaluation.[3]

The general problem consists of selecting a subset of $\mathcal{X} \times \mathcal{M}$ that is evaluated (with cost under $B$) such that the evaluation result on the subset (e.g. model ranking) is the same as if we evaluated on the full $\mathcal{X} \times \mathcal{M}$.

Instead of selecting everything at the same time, we turn the problem into a sequential decision-making problem so that we can use the already-collected evaluations to inform the next selection. At each step $i$, we select a particular model $m_i \in \mathcal{M}$, for which we evaluate an item.

To make the problem tractable, we further split this problem in two: (1) selecting which model to evaluate, and (2) selecting which item to evaluate. The latter is the focus of existing works (Section 2) and gives us some item ordering $\prec$ that orders $\mathcal{X}$ from most to least informative. Therefore by selecting a model, we evaluate it on the first item it has not been evaluated on yet (Figure 2).[4] Formally, we tackle the problem by learning a selection policy $\pi$:

$$\pi : \bigcup_{i \in \mathbb{N}_0} (\mathcal{X} \times \mathcal{M})^i \to \mathcal{M} \quad (1)$$

[2] See Appendix A for a more general problem with joint model and evaluation item selection that leads to our constrained version through a series of observations.

[3] We assume the human evaluation cost is primarily given by the input item and independent of the output (Kocmi et al., 2024a) and all inference costs are negligible in comparison.

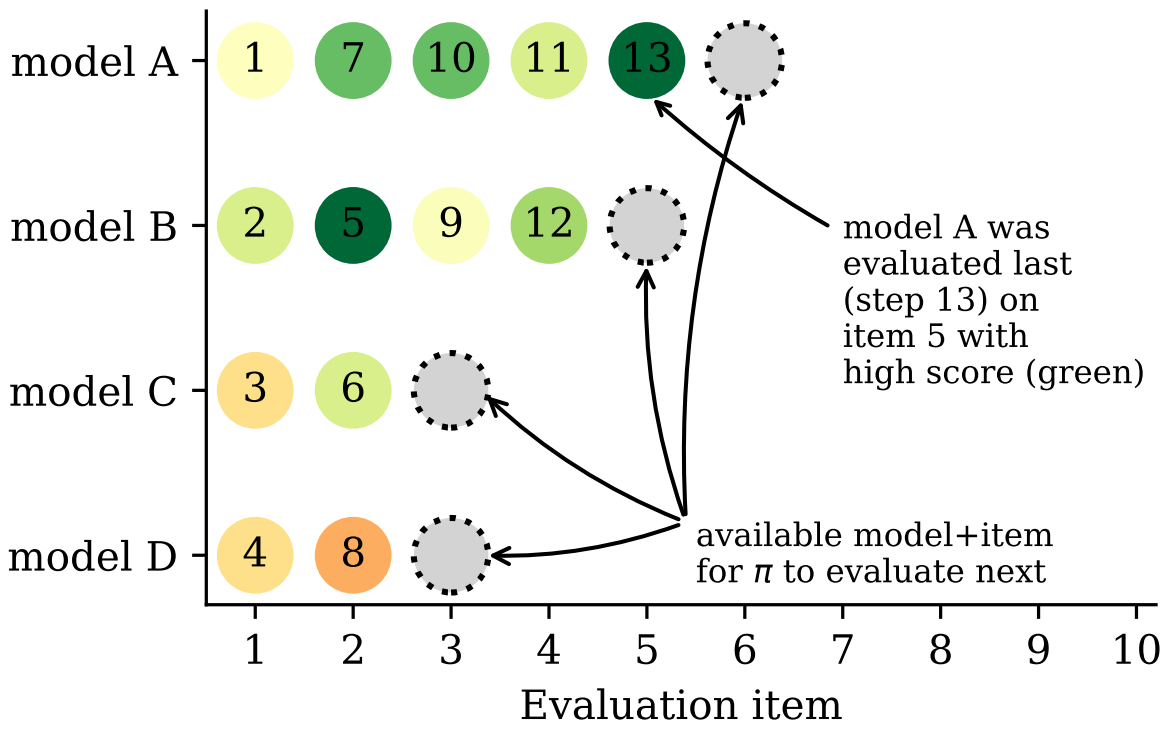

Figure 2: Constrained evaluation allocation problem. Based on evaluation history, we select which model is going to be evaluated next on the next item it has not been evaluated on so far. Numbers indicate at which step was the model+item combination evaluated.

Given a selected model $m_i$, we choose the first item where the model is not evaluated yet:

$$x = \min_{\prec}\{x \,|\, x \in \mathcal{X}, \langle x, \pi(R^\pi_{i-1})\rangle \notin R_{i-1}\} \quad (2)$$

At each step $i$ we have access to the previous evaluation results $R^\pi_{i-1}$:

$$R^\pi_i = R^\pi_{i-1} \cup \{\langle x, \pi(R^\pi_{i-1})\rangle\} \quad (3)$$

The stopping criteria is given by budget. We therefore find the last selection step within budget:

$$T_\pi = \max\left\{ 1 \le i \le |\mathcal{X}| \,\middle|\, \sum_{\langle x,m\rangle \in R^\pi_i} \text{cost}(x) \le B \right\} (4)$$

Ultimately, we wish to obtain the policy that optimizes some objective of the final evaluation result under the budget.

$$\pi^* = \text{argmax}_\pi \text{ Objective}\left(R^\pi_{T_\pi}\right) \quad (5)$$

To make the above problem tractable, we now discuss a set of simplifying assumptions on $\prec$ and $\text{cost}(x)$.

**Assumption 1: Unbiased item ordering.** For policy $\pi$ we might need to estimate current model performance *during* evaluation, which can be done by simply taking the mean across all evaluated items. Biased item orderings, such as ordered based on decreasing difficulty makes our mean estimator biased: models evaluated on fewer items can be at a disadvantage compared to others as they have been exposed only to difficult items. For this reason, we assume that the item ordering $\prec$ is unbiased. See Lemma 3 for elaboration on biased estimators.

Later, we relax this assumption by estimators that take item difficulty into account, allowing for arbitrary item orderings (e.g. prioritizing difficult items; Proietti et al., 2025; Zouhar et al., 2025).

**Assumption 2: Constant cost.** We assume that evaluation cost is constant for all items $\forall x \in \mathcal{X} : \text{cost}(x) = 1$. Note that while this is a strong assumption as the cost of evaluating longer outputs might be more than that of shorter outputs, again our estimators might become biased if we prioritize models for whom the next evaluation item is e.g. cheap. Later we will relax this assumption by incorporating $\text{cost}(x)$ into that item ordering $\prec$.

### 3.2 Objective of policy $\pi$

Next we will present our objective for learning the policy $\pi$. This depends on the evaluation goals: Do we care only about finding the top-1 model, or the ranking among top-3 models, or ranking of all models with a particular focus on the top ones?

We consider the correlation between the model ranking obtained from $R^\pi_{T_\pi}$ against the "true model ranking" obtained from the entire set of possible evaluations $\mathcal{X} \times \mathcal{M}$. A typical approach would be to use ranking correlation such as Kendall's $\tau$. However, common ranking correlations equally weigh ranking of top and low-performing models, which misses our focus on top models. Thus, we turn to **weighted Kendall's $\tau_\omega$** (Vigna, 2015) in which exchanges of high weight are more influential than exchanges of low weight. The weighting function can be arbitrarily set to the inverse rank of the model $m$, $\frac{1}{\text{rank}_m}$, which assigns higher weights to top-ranking models, or a step-function (considering only top 3 models or top 1 model), or a more skewed version of inverse rank, $\frac{1}{\text{rank}^2_m}$. We choose the latter for the main experiments of this paper for high weight assigned to top-1 model being correct, but include many other $\omega$ choices in Appendix C.3.

Given true average model scores $\mu$ and the average model scores on the evaluated items $\hat{\mu}$ we define the weighted Kendall tau:

[4] See Appendix A for reduction elaboration.

$$\tau_\omega(\hat{\mu}, \mu) \stackrel{\text{def}}{=} \frac{\sum_{i \neq j} \omega_i \omega_j \cdot \text{sign}(\mu_i - \mu_j) \, \text{sign}(\hat{\mu}_i - \hat{\mu}_j)}{\sum_{i \neq j} \omega_i \omega_j} \quad (6)$$

where, sign : $\mathbb{R} \rightarrow \{-1, 0, 1\}$.

We discuss other evaluation objectives in Appendix C.2, such as average payoff, top-$k$ accuracy, stability of ranking, and statistical discriminability.

### 3.3 Problem Hardness

We now show that even with both Assumptions 1 and 2 and with oracle access to evaluations, the optimization problem is NP hard, which prevents us from attaining optimal policy $\pi^*$.

**Theorem 1** (Evaluation allocation is NP hard, proof in Appendix B): Even with oracle access to $\mathcal{M} \times \mathcal{X}$, finding a subset $R^\star$ such that $|R^\star| \leq B$ and the model mean estimates based on $R^\star$ maximize $\tau_\omega(\hat{\mu}, \mu)$ is NP hard for arbitrary $\omega$.

### 3.4 Unconstrained Item Ordering

Next, we consider lifting Assumptions 1 and 2 which mandate a random unbiased evaluation item ordering, no annotator drift, and constant cost. Naively with mean-based estimators $\hat{\mu}_m$, this can lead to situations where we compare models (their $\mu_m$) where for some $m$, $\hat{\mu}_m$ is computed on many items and for some where it is computed only on the first few items, which might happen to be unfairly more difficult, for example.

To this end, we adopt methods which smartly estimate $\hat{\mu}_m$ by first predicting missing evaluations (Figure 3) and then computing mean on the same set of items for all models. Similar methods exist, known as Item Response Theory or EASL (Balkır et al., 2026; Sakaguchi and Van Durme, 2018),

Thus, given a set of previous evaluations, we estimate $r_{x,m}$ even for $\langle x, m \rangle \notin R$. We do so by assuming that the observed score is given by an additive model (Cronbach et al., 1972):

$$r_{x,m} = q_m + d_x + \varepsilon_{x,m} \quad (7)$$

where $q_m$ is the true latent quality of model $m$, $d_x \sim \mathcal{N}(0, \sigma_d^2)$ is the inherent difficulty of item $x$, and $\varepsilon_{x,m} \sim \mathcal{N}(0, \sigma_\varepsilon^2)$ is independent observation noise. We fit the parameters $\boldsymbol{q}$ and $\boldsymbol{d}$ based on the existing observations $R$ using the objective below:

$$\mathcal{L} = \|r_{x,m} - \left(\hat{q}_m + \hat{d}_x\right)\|^2 \quad (8)$$

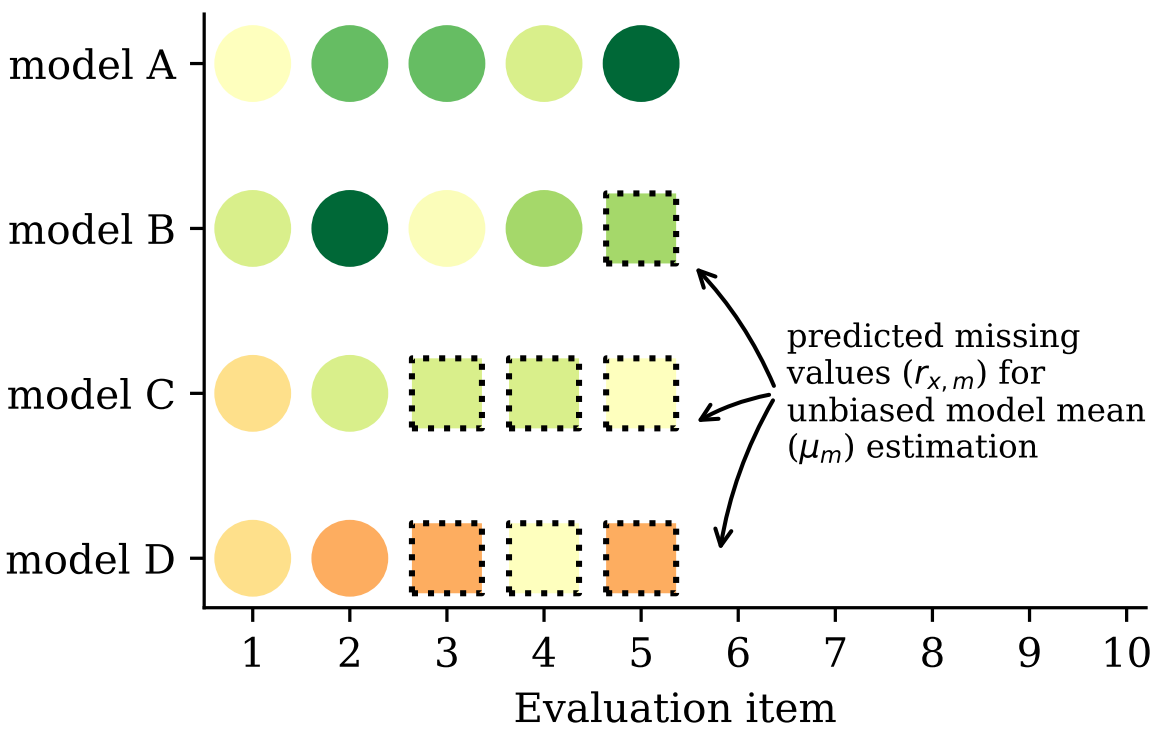

Figure 3: "Missing" values are predicted with a linear model such that the model mean estimates $\mu_m$ are computed on the same set of items, which might vary in difficulty. Extension of Figure 2.

and then even for unseen model $m$'s evaluation on item $x$, we predict:

$$\hat{r}_{x,m} = \hat{q}_m + \hat{d}_x \quad (9)$$

Then, when we compute $\hat{\mu}_m$ or $\text{rank}_m$ we do so based on the full set $\{\hat{r}_{x,m}\}_{\mathcal{X} \times \mathcal{M}}$.

By removing Assumptions 1 and 2, we can now arbitrarily modify the ordering of evaluation items. Formally, each item has an assigned $\text{Utility}(x)$, which can be, for example, its difficulty, discriminability, diversity, or variance (Proietti et al., 2025; Zouhar et al., 2025). We use this utility to order evaluation items $\mathcal{X}$, so $x \prec x' \Leftrightarrow \text{Utility}(x) < \text{Utility}(x')$.

We make the item ordering cost-aware by considering the $\text{Utility}(x)$ per unit of cost: $\frac{\text{Utility}(x)}{\text{cost}(x)}$. In general, cost-aware evaluation item selection is NP-complete and can only be approximated (Zouhar et al., 2025). However, in our case all items cost much less than the total budget ($\max \text{cost}(x) \ll B$), which makes the above density-based heuristic asymptotically optimal (Dantzig, 1957).

## 4 Algorithms

Our optimization (Equation 5) is similar to best-arm identification problem in multi-armed bandits, where the objective is to identify the top arm with high probability rather than maximizing cumulative reward (Audibert and Bubeck, 2010; Gupta et al., 2021). Pulling an arm, corresponding to a model, means evaluating said model on the next item, which comes at a price but reveals more information about the model ranking. We discuss several algorithmic choices, their optimality under uncertainty, but also the general difficulty of finding the optimal allocation, as demonstrated by NP-hardness of oracle solution to the problem.

Each model $m \in \mathcal{M}$ here corresponds to an arm; pulling arm $m$ reveals its score on the next available item $x$ in the evaluation item queue. Unlike the classic regret minimization objective, which seeks to maximize the cumulative reward, our setting is one of pure exploration. We seek to allocate the fixed budget $B$ to find the correct model ranking with weights according to $\omega$, as in Equation 6. This approach affords greater flexibility in focusing evaluation on top-tier models, but necessitates a warmup phase, where all models are first evaluated on first $C$ items, to stabilize initial estimates of $\hat{\mu}_m$.

The simplest and most common baseline is to randomly draw the next model to evaluate from $\mathcal{M}$ at each turn. In practice, we implement this baseline where every model is evaluated on an identical number of items, $\frac{B}{|\mathcal{M}|}$. While this provides an unbiased estimate of the global ranking, it wastes budget on clearly inferior models that may not require further discrimination.

### 4.1 Algorithm: Weighted Sampling

We consider three versions of weighted sampling corresponding to various strategies for selecting the next model at each step (see Algorithm 1).
1. **$\varepsilon$-greedy sampling**, which selects the current top model and occasionally (with probability $\varepsilon$), the other models:

$$\omega_m = 1\text{-}\varepsilon \text{ if } \text{rank}_m\text{=}1 \text{ else } \varepsilon/|\mathcal{M}| \quad (10)$$

2. Sampling from the **Boltzmann distribution** over the current score averages, a standard exploration strategy in reinforcement learning (Sutton and Barto, 2018):

$$\omega_m = \exp(\hat{\mu}_m/\text{temp.}) \quad (11)$$

3. Sampling proportional to the inverse of the **model rank** to the power $k$, which draws on selection mechanisms from evolutionary computation (Baker, 1985) and is not sensitive to the scale of $\mu$:

$$\omega_m = 1/\text{rank}_m^k \quad (12)$$

Notably, this intentionally bears resemblance to weights in optimization objective (Equation 6).

Additionally, we also include **upper confidence bound**, which is a standard algorithm that balances the exploration-exploitation trade-off by selecting models that maximize the UCB score (Auer et al., 2002):

$$\arg\max_m \hat{\mu}_m + \gamma\sqrt{\ln|R| \;/|R_m|} \quad (13)$$

The UCB score of each model consists of the empirical mean estimate of the model average $\hat{\mu}_m$ (exploitation) and a confidence term accounting for the number of times the model has been evaluated (exploration).

A desirable property of allocation algorithms is *consistency*, where with enough samples, we arrive at the correct model ranking ($\tau_\omega = 1$). Notably, UCB is not consistent while other algorithms presented above are (shown in Theorem 2).

**Theorem 2** (Consistency of allocation algorithms, proof in Appendix B): Let $\mathcal{M}$ be a finite set of models and assumptions 1 and 2 hold. Let $\pi$ be an allocation policy such that for any step $t$, the selection probability $P_t(m)$ for every model $m$ satisfies $P_t(m) \geq \zeta$ for some fixed constant $\zeta > 0$. Then, the estimated ranking converges to the true ranking as $B \to \infty$.

We now show that when our objective (Equation 6) is weighted according to $\omega$ and data is homoscedastic (Appendix D), then sampling according to $\sqrt{\omega}$ is optimal (Theorem 3). In our case for the weighted $\tau_\omega$, we use $\frac{1}{\text{rank}_m^2}$ which corresponds to optimality of weighted sampling that samples according to $\frac{1}{\text{rank}_m}$. Conversely, this makes uniform sampling optimal when the weights $\omega$ are constant because $\omega = \sqrt{\omega} = 1$.

**Theorem 3** (Optimality of weighted sampling for weighted $\tau_\omega$, proof in Appendix B): Assume the estimated model score averages $\hat{\mu}_m$ follows an independent Gaussian distribution $\mathcal{N}\left(\mu_m, \frac{\sigma^2}{|R_m|}\right)$. Sampling according to $\sqrt{\omega_m}$ is the optimal strategy for minimizing the weighted (according to $\omega_m$) ranking uncertainty.

### 4.2 Algorithm: Confusion minimization

We now present an alternative approach to weighted sampling. The optimality of weighted sampling algorithm relies on the assumption of homoscedasticity (constant variance for all models in scores per item), which is violated in practice (Appendix D). Furthermore, weighted sampling does not consider uncertainty of ranking and the distances between model mean estimates (corresponding to probability of two neighboring models being swapped) are not taken into account. For example, even if the top model's mean is estimated to be much higher than the one of the runner-up, $\hat{\mu}_1 \gg \hat{\mu}_2$ is, we still waste evaluation effort on obtaining a more precise $\hat{\mu}_1$. This is despite us being already confident that $m_1$ is better than $m_2$.

In response, we focus on the pairwise rankings in the $\tau_\omega$ objective. However, directly optimizing Equation 6 is impossible without knowing the true ranking. Thus, as a proxy, we propose to maximize the weighted certainty of model ranking.

At each step in policy $\pi$, we have to choose a model $a$ for which we evaluate one new evaluation item, which improves its $\hat{\mu}_a$ estimate. To best use this choice, at each step we select model $a$ for which across all other models $b \in \mathcal{M} \setminus \{a\}$ the weighted increase in ranking corectness after adding an extra evaluation item is maximized:

$$\omega_a \cdot \sum_{b \in \mathcal{M} \setminus \{a\}} \omega_b \cdot \left( P^+_{a,b} - P_{a,b} \right) \quad (14)$$

$$\begin{aligned} P_{a,b} &= P(\hat{\mu}_a < \hat{\mu}_b \Leftrightarrow \mu_a < \mu_b) \\ P^+_{a,b} &= P(\hat{\mu}^+_a < \hat{\mu}_b \Leftrightarrow \mu_a < \mu_b) \end{aligned} \quad (15)$$

We first estimate $P_{a,b}$ (probabilty that estimated ranking is the same as true ranking) and make the assumption that the probability of correctly ranking models $a$ and $b$ is the maximum between the probability that $\hat{\mu}_a > \hat{\mu}_b$ and the probability of $\hat{\mu}_a < \hat{\mu}_b$. This corresponds to our certainty of this pairwise ranking in any direction:

$$\begin{aligned} &P(\hat{\mu}_a < \hat{\mu}_b \Leftrightarrow \mu_a < \mu_b) \\ &\approx \max(P(\hat{\mu}_a < \hat{\mu}_b), P(\hat{\mu}_a > \hat{\mu}_b)) \end{aligned} \quad (16)$$

We choose to model $P(\hat{\mu}_a < \hat{\mu}_b)$ using normal distribution (see verification of this assumption in Appendix D, $\Phi$ is the cummulative distribution function of the standard normal distribution):

$$\begin{aligned} &\max(P(\hat{\mu}_a < \hat{\mu}_b), P(\hat{\mu}_a > \hat{\mu}_b)) \\ &= \Phi\left( \frac{|\hat{\mu}_a - \hat{\mu}_b|}{\sqrt{\frac{\sigma_a^2}{|R_a|} + \frac{\sigma_b^2}{|R_b|}}} \right) \end{aligned} \quad (17)$$

Here, $R_a$ and $R_b$ are sets of evaluated items for models $a$ and $b$ at this particular timestep.

Computing $P^+_{a,b}$ in Equation 15 is more difficult. We do not know yet know the new evaluation item, so the best guess is that the new mean remains the same, $\hat{\mu}^+_a = \hat{\mu}_a$. However, we know that it will lower the variance because the sample size of model $a$'s collected evaluations $|R_a|$ will contain one extra item.

$$\begin{aligned} &P(\hat{\mu}^+_a < \hat{\mu}_b \Leftrightarrow \mu_a < \mu_b) \\ &\approx \max(P(\hat{\mu}^+_a < \hat{\mu}_b), P(\hat{\mu}^+_a > \hat{\mu}_b)) \end{aligned} \quad (18)$$

$$= \Phi\left( \frac{|\hat{\mu}_a - \hat{\mu}_b|}{\sqrt{\frac{\sigma_a^2}{|R_a|+1} + \frac{\sigma_b^2}{|R_b|}}} \right) \quad (19)$$

The algorithm then turns into selecting such a model that maximizes Equation 14 (details in Algorithm 2). A key advantage of confusion minimization is that it allows late addition of new models, which will be selected more frequently thanks to high predicted reduction in uncertainty.

### 4.3 Algorithm: Greedy oracle

To quantify the performance of our allocation algorithms, we wish to establish an upper bound. Because it is effectively impossible to find the optimal allocation, we turn to greedy approximation[5] which at each step scans one choice ahead and selects the model that maximizes the weighted ranking correlation against the true ranking:

$$\arg\max_{m \in \mathcal{M}} \tau_\omega(R \cup \{\langle x_m, m\rangle\}, \mathcal{X} \times \mathcal{M}) \quad (20)$$

$$x_m = \min_{\prec}\{x \,|\, x \in \mathcal{X}, \langle x, m\rangle \notin R\} \quad (21)$$

This is an unrealistic algorithm in practice due to the oracle access to $\mathcal{X} \times \mathcal{M}$ in Equation 20. However, it serves as an approximation of the upper bound.

**Other algorithms.** In Appendix C.1 we describe other algorithms, such as scheduled discarding of significantly worst models or minimizing variance). However, they are either have undesireable properties or underperform.

## 5 Experiments

We evaluate the proposed dynamic policies by using existing annotation campaigns from WMT General Translation shared tasks (Kocmi et al., 2023, 2024a, 2025), which typically has $|\mathcal{M}| \approx 20$ competing models and $|\mathcal{X}| \approx 400$ evaluation items across ∼20 language pairs every year. The score distribution per item varies greatly (Appendix D), which makes finding the top model difficult. Our simulation-based analysis allows us to compare our policies against a "gold standard" ranking derived from the dataset of exhaustive evaluations for all model-item pairs while controlling the annotation budget.

### 5.1 Case Study on Translation Evaluation

We use the data collected by the WMT General Translation Shared Task between 2023 and 2025 (Kocmi et al., 2023, 2024a, 2025) across 37 cam-

[5] The proposed greedy oracle solution is high-performing in practice. However, the objective function is not submodular and it is possible to construct an adversarial case such that the greedy solution is arbitrarily worse than the optimum.

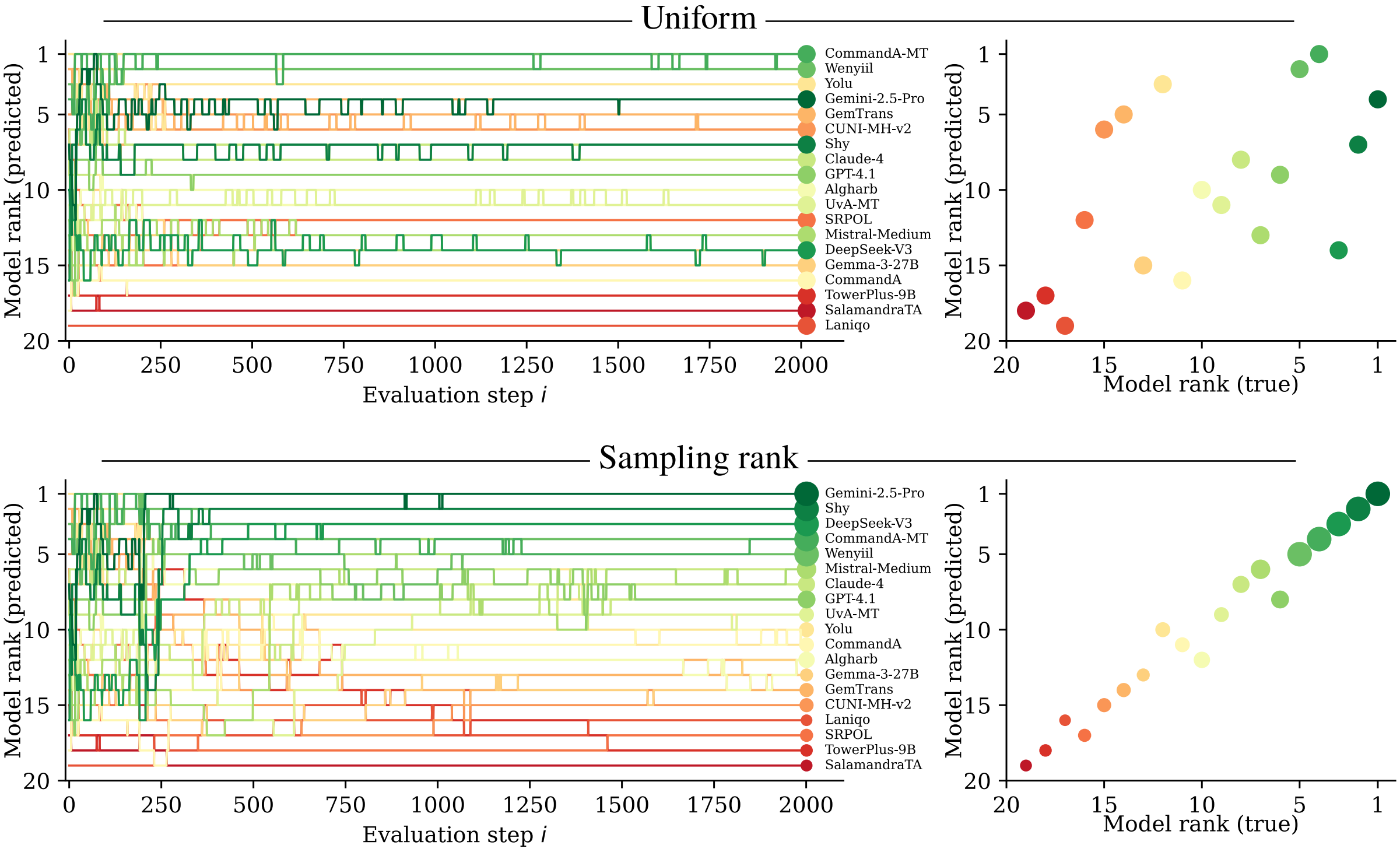

Figure 4: Timeline of evaluation based on WMT25 English→Czech campaign. Size of ● corresponds to number of evaluated items and its color to true rank. Uniform sampling balances equally across all models while rank-based sampling prioritizes top-performing models, which leads to fewer mistakes in ranking top models.

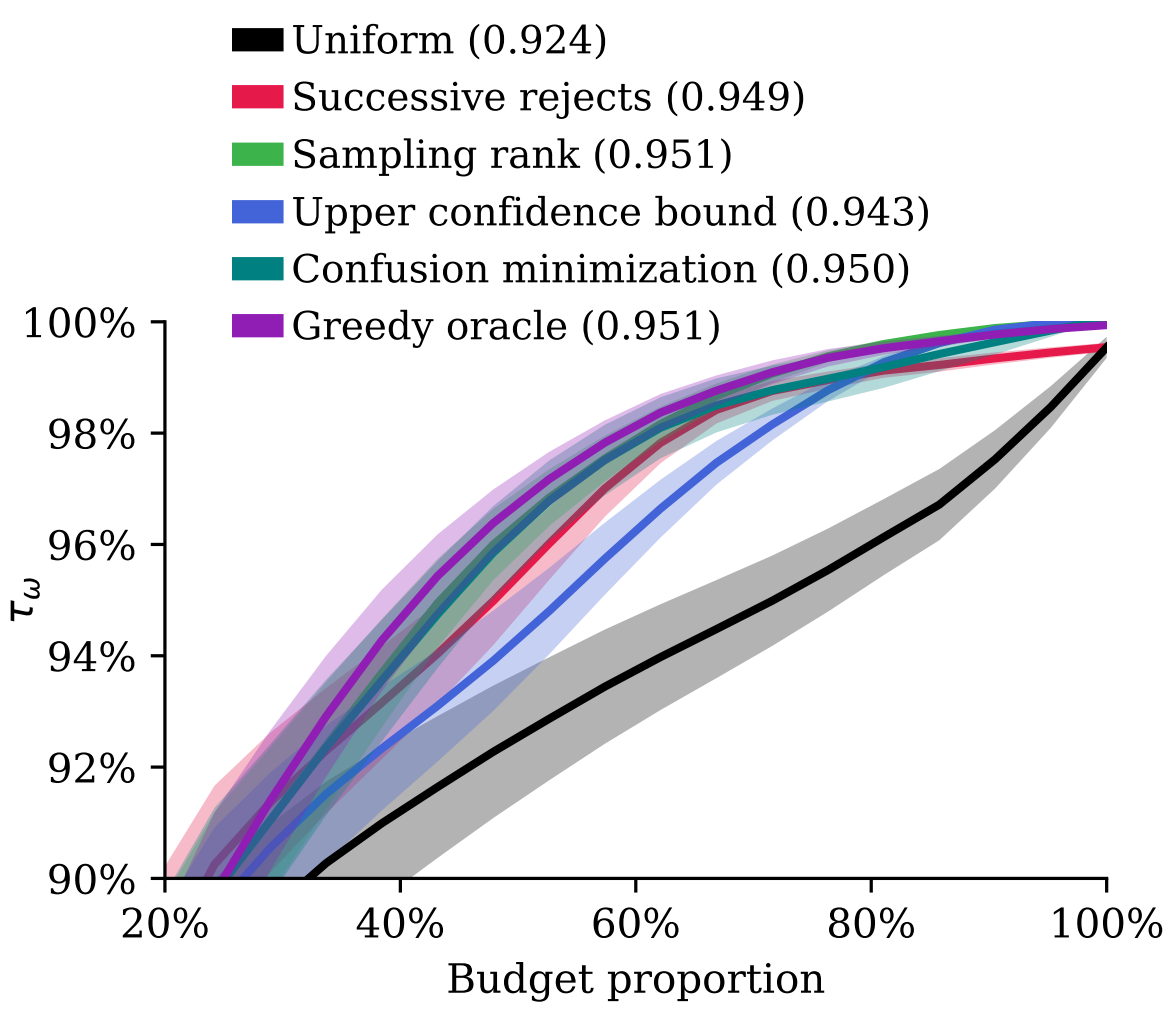

Figure 5: Ranking $\tau_\omega$ (average) across varying budgets (proportion of all item+model evaluations in the data), averaged across all languages and 100 seeds. Shaded areas correspond to 95% t-distribution confidence interval. Numbers in legend show areas under curves (i.e. average $\tau_\omega$).

paigns (language pair + year), with $|\mathcal{X}|$=570 items and $|\mathcal{M}|$=15 models on average, with human evaluation (290k evaluations in total) done with human annotation protocols described in (Kocmi et al., 2024b; Freitag et al., 2021), the outcome of which is a score in $[0, 1]$ for each input item and model pair.

All algorithms use coldstart $C = 5$ items evaluated across all models. For algorithms we also use temp. $= 1, \varepsilon = 50\%$, and $\gamma = \sqrt{2}$ for UCB as defaults that balance exploration and exploitation.

**Results zoomed-in.** We first show an illustrative run of our simulated evaluation on the WMT25 English→Czech campaign done with uniform selection and rank-based sampling in Figure 4. Under the uniform policy, the rankings for mid- to top-performing models are volatile, which persists well into the later evaluation steps due to much of the budget being spent over inferior models. Conversely, the rank-based sampling policy focuses on top-performing models. The size of the dots on the right show that each model is allocated the same number of evaluations under the uniform policy while it follows $\frac{1}{\text{rank}}$ for rank-based sampling allowing more evaluations for higher ranked models. As a result, the ranking based on the uniform sampling is further away from the true model ranking.

**Results across budgets.** In Figure 5, we show the results across all campaigns and varying budget. Naturally, the higher the budget, the better the final ranking. As shown, our proposed algorithms

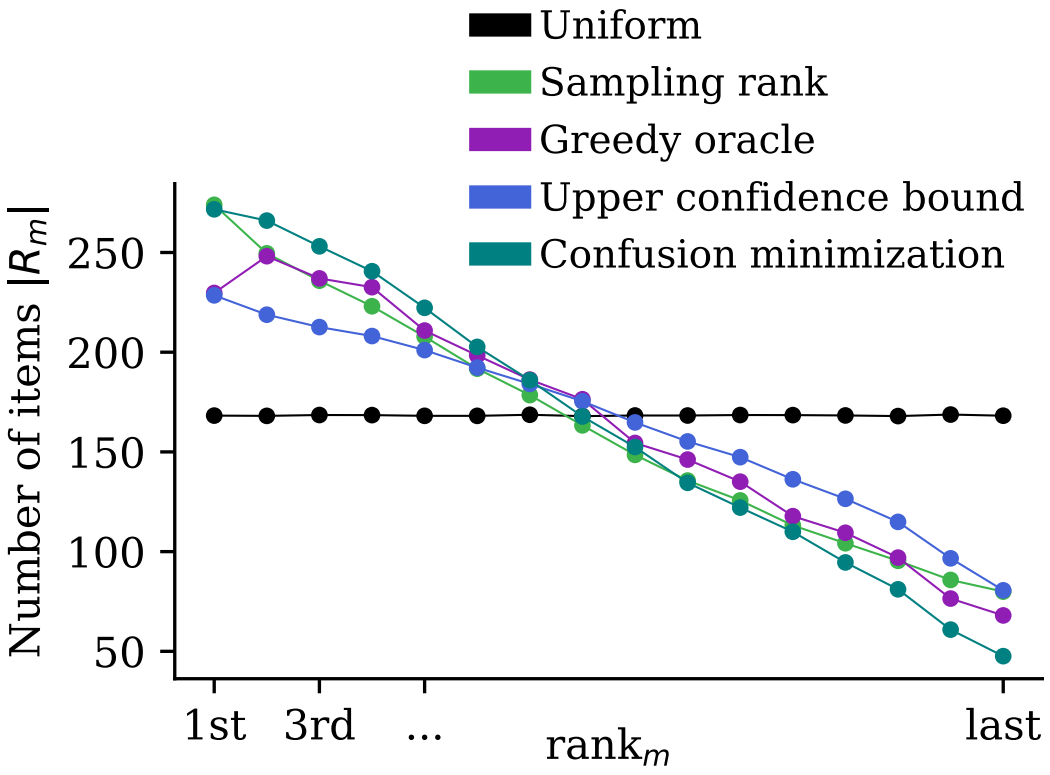

Figure 6: Resulting allocation of evaluation items from various algorithms with respect to model rank. Our algorithms are close to oracle allocation.

outperform uniform sampling and approach the greedy oracle selection.

Other algorithms are included in Appendix C.1. In Appendix C we show that non-uniform sampling also leads to higher stability across stochasticity in data.

**Allocation of evaluation effort.** We examine the distribution of the allocated budget that the weighted sampling and other top-performing algorithms assigned to models of various ranks. Figure 6 shows that this distribution closely follows the distribution by greedy oracle.

**Unconstrained item ordering.** We now use smart item ordering, which requires lifting assumptions 1 and 2 (Section 3.4). Specifically, we order the evaluation items based on their difficulty[6] (easy- or hard-first), and also normalize by the evaluation cost estimated through the annotation time. The results are shown in Table 1. As per Lemma 3 mean-based estimator of model performance make the non-uniform algorithms perform worse than with random ordering. By applying the linear estimators from Equation 7, the bandit algorithms are also able to benefit from the smart item ordering.

| | | Rand. | Easy | Hard | Hard /cost |
|---|---|---|---|---|---|
| | | $\tau_w$ | $\tau_w$ | $\tau_w$ | $\tau_w$ |
| Mean estimator | Uniform | 0.924 | 0.873 | 0.928 | 0.942 |
| | Successive rejects | 0.949 | 0.749 | 0.948 | 0.930 |
| | Sampling rank | 0.951 | 0.849 | 0.939 | 0.949 |
| | Sampling rank$^2$ | 0.948 | 0.811 | 0.928 | 0.936 |
| | Sampling $\varepsilon$-greedy | 0.947 | 0.824 | 0.935 | 0.939 |
| | Sampling Bolzmann | 0.943 | 0.894 | 0.942 | 0.955 |
| | Upper confidence bound | 0.943 | 0.896 | 0.948 | 0.960 |
| | Confusion minimization | 0.950 | 0.803 | 0.949 | 0.959 |
| | Greedy oracle | 0.951 | 0.649 | 0.944 | 0.942 |
| Linear estimator | Uniform | 0.924 | 0.873 | 0.928 | 0.942 |
| | Successive rejects | 0.950 | 0.912 | 0.949 | 0.956 |
| | Sampling rank | 0.950 | 0.903 | 0.951 | 0.962 |
| | Sampling rank$^2$ | 0.942 | 0.879 | 0.935 | 0.947 |
| | Sampling $\varepsilon$-greedy | 0.942 | 0.891 | 0.942 | 0.949 |
| | Sampling Bolzmann | 0.944 | 0.894 | 0.946 | 0.958 |
| | Upper confidence bound | 0.943 | 0.902 | 0.946 | 0.960 |
| | Greedy oracle | 0.951 | 0.897 | 0.952 | 0.964 |

Table 1: Average $\tau_\omega$ for selection with biased ordering of evaluation items (easy-first, difficult-first, cost-normalized difficult-first). Model estimates are done either with means or a linear additive model.

## 6 Conclusion

Standard uniform evaluation protocols inefficiently allocate effort to inferior models, which hinders large-scale evaluation. We formalize human evaluation as best-arm identification within a multi-armed bandit framework with correlated arms and prove the optimality of proposed algorithms under certain conditions. In contrast, our dynamic evaluation allocation trades off global model ranking precision for ranking precision among top models. This makes evaluations faster, cheaper and more aligned with our objectives.

Intended use-cases include: (1) Benchmarking campaigns with focus on state-of-the-art models such as WMT (Kocmi et al., 2025) or IWSLT (Abdulmumin et al., 2025).[7] (2) Tournament settings where evaluations must be routed to high-uncertainty, top-tier candidates, such as Chatbot Arena (Chiang et al., 2024) and (3) Rapid model checkpoint or configuration selection, including scenarios even automatic metrics (e.g. LLM-as-a-judge) become expensive at scale.

**Recommendations:** We provide recommendations to practitioners conducting evaluations. First, evaluation goals should be established: Are you interested in all models equally or the top ones? Then, the evaluation goals should be quantified into $\omega$ (e.g. $\omega = 1/\text{rank}_m^2$) and models sampled according to $\sqrt{\omega}$. For translation and multilingual tasks, it is already implemented in Pearmut (Zouhar and Kocmi, 2026) annotation platform.

[6] Average automatics metric scores by MetricX (Juraska et al., 2023, 2024, 2025).

[7] For example, the WMT 2026 (Kocmi et al., 2026) general machine translation shared task already adopted dynamic sampling for human evaluation live at scale.

## Limitations

Our experiments, due to their scale, are simulated on existing evaluation data. This creates a natural cap on how many times we can evaluate a particular model (see Figure 4) and obscures potential behaviors in scenarios where the item pool $\mathcal{X}$ is vastly larger than the budget $B$, which prevents full convergence of the mean estimator. Larger data scale are economically infeasible and we address this constraint by synthetic scale in Appendix C.4.

Most of the proposed methods harbor the practical risk of annotation drift: as the phases progress, annotators are exposed to an increasingly high-quality pool of models (less so with methods which we show are *consistent*, Theorem 2), which may cause them to recalibrate their internal scoring metrics. We delegate this concern to the design of more objective human annotation protocols.

## Ethics statement

*The Leaderboard Illusion* (Singh et al., 2025) demonstrates that sampling asymmetries in live platforms can lead to gaming rankings via private testing. Adaptive allocation policies in public arenas therefore requires full transparency constraints to safeguard evaluation integrity.

No new data has been collected for this study and we foresee no further ethical concerns.

## Acknowledgements

Vilém Zouhar gratefully acknowledges the support of the Google PhD Fellowship. Ondřej Bojar would like to acknowledge the support of the grant CZ.02.01.01/00/23_020/ 0008518 (Jazykověda, umělá inteligence a jazykové a řečové technologie: od výzkumu k aplikacím). This research has been funded in part by a Swiss National Science Foundation award (project 201009) and a Responsible AI grant by the Haslerstiftung.

# A General Problem Statement

We describe a general form of the dynamic evaluation problem, that jointly selects both which *models* and *items* to evaluate. We then constrain it through two practical observations to only model selection for evaluation, which is the focus of this work and is used in the main paper.

**Problem (general).** Let $\mathcal{M}$ be a set of models, $\mathcal{X}$ a set of atomic evaluation items, and budget $B \in \mathbb{R}^+$. Given the output of model $m$ on item $x$, we have its evaluation $r_{x,m} \in [0, 1]$ and also the cost of this evaluation, $\text{cost}(x) \in \mathbb{R}^+$.[8] At each step $i$ we select a particular item $x \in \mathcal{X}$ and model combination $m \in \mathcal{M}$, which is stored $R_i = R_{i-1} \cup \{\langle x, m\rangle\}, R_0 = \{\}$. During the selection at step $i$, we have access to previous results $R_{i-1}$. Formally, we define the allocation policy $\pi$ that maps from history to the next selection:

$$\begin{aligned} \pi &= \bigcup_{i \in \mathbb{N}_0} (\mathcal{X} \times \mathcal{M})^i \to \mathcal{X} \times \mathcal{M} \\ R_i^\pi &= R_{i-1}^\pi \cup \{\pi(R_{i-1}^\pi)\} \end{aligned} \tag{22}$$

We find the last selection step within budget:

$$T_\pi = \max\left\{ 1 \leq t \leq |\mathcal{X}| \;\middle|\; \sum_{\langle x,m\rangle \in R_t^\pi} \text{cost}(x) \leq B \right\} \tag{23}$$

Ultimately, we wish to obtain the policy that optimizes the objective of the final evaluation result under the budget.

$$\pi^* = \text{argmax}_\pi \ \text{Objective}\Big(R^\pi_{T_\pi}\Big) \tag{24}$$

**Observation 1: Not repeating annotations.** In the current statement it is possible to select the same evaluation item for the same model multiple times. This improves the estimate at a particular evaluation item due to the reduced annotation noise. However, this is suboptimal for estimating the model mean due to the wasted information in covariance of the duplicate items. See Lemma 1 for a formal proof. Empirically with real annotations this has also been shown by Riley et al. (2025) and we can thus safely avoid duplicate model-item selections.

**Observation 2: Item overlap between models.** Evaluation noise stems not only from annotator disagreement but from the variance in difficulty across evaluation items. When comparing two models on different sets of items, we introduce a confounder. For evaluation practitioners, it is more beneficial to have models evaluated on the same set items so that they can be more directly compared. While we still wish to allow for the better models to be evaluated more often, we wish to prevent models having the same number of evaluations but on different items.

Importantly, under an additive error model, maximizing overlap minimizes the variance of the estimator for the difference in model quality (see Lemma 2). Practically, if we maximize the overlap between the two models, we can use paired statistical tests in post-hoc comparison rather than independent tests. Except for special cases of active evaluation, the vast majority of evaluation benchmarks fulfills this criterion by evaluating all models on the same set of items.

As a corollary, if further for all models $m, m' : |R_m| - |R_{m'}| \leq 1$ then the problem statement is equivalent to having a fixed item ordering and the policy $\pi$ only chooses which model to evaluate next. Effectively this reduces the dimensionality of the bandit action space from $|\mathcal{X}| \times |\mathcal{M}|$ to $|\mathcal{M}|$. We accept this simplification and practically we order $\mathcal{X}$ based on some ordering $\prec$.

[8] The human evaluation cost is primarily given by the input item and independent of the output (Kocmi et al., 2024a). We assume that all inference costs are negligible in comparison.

## B Proofs and Formalization Details

**Lemma 1** (Non-repeating annotations): The variance of the sample mean estimator $\hat{\mu}_m$ of a model $m \in \mathcal{M}$ is strictly minimized when the selection policy $\pi$ selects distinct items.

**Proof**: Let $\hat{\mu} = \frac{1}{B}\sum_{i=1}^{B} r_i$ estimate average model score under the additive model $r_i = q_m + d_i + \varepsilon_i$. The variance of this sum is:

$$\mathrm{Var}(\hat{\mu}) = \frac{1}{B^2}\left(\sum_{i=1}^{B} \mathrm{Var}(r_i) + \sum_{i \neq j} \mathrm{Cov}(r_i, r_j)\right) \tag{25}$$

where the variance of any single observation is constant: $\mathrm{Var}(r_i) = \sigma_d^2 + \sigma_\varepsilon^2$.

Case 1: (distinct items): Evaluations are independent. The covariance between any two scores is zero.

$$\mathrm{Var}(\hat{\mu}_{\text{distinct}}) = \frac{1}{B^2}(B(\sigma_d^2 + \sigma_\varepsilon^2)) = \frac{\sigma_d^2 + \sigma_\varepsilon^2}{B} \tag{26}$$

Case 2 (repeated items): Suppose an item $x$ is evaluated twice, yielding $r_1$ and $r_2$. They share the identical item difficulty $d_x$. Their covariance is strictly positive:

$$\mathrm{Cov}(r_1, r_2) = \mathrm{Cov}(q_m + d_x + \varepsilon_1, q_m + d_x + \varepsilon_2) = \mathrm{Var}(d_x) = \sigma_d^2 > 0 \tag{27}$$

As a result:

$$\mathrm{Var}(\hat{\mu}_{\text{repeat}}) = \mathrm{Var}(\hat{\mu}_{\text{repeat}}) + \mathrm{Cov}(r_1, r_2) > \mathrm{Var}(\hat{\mu}_{\text{distinct}}) \tag{28}$$

■

**Lemma 2** (Item overlap between models): Let $\hat{\Delta}_{m,m'}$ be the estimator for the quality difference between two models $m$ and $m'$, given evaluation sets $R_m$ and $R_{m'}$ of fixed sizes. Under an additive random effect model where item difficulties are independent and identically distributed, the variance $\mathrm{Var}\left(\hat{\Delta}_{m,m'}\right)$ is strictly minimized when the intersection $|R_m \cap R_{m'}|$ is maximized.

**Proof**: We model the score $r_{x,m}$ of a model $m$ on item $x$ using a standard additive error decomposition:

$$r_{x,m} = q_m + d_x + \varepsilon_{x,m} \tag{29}$$

where $q_m$ is the true latent quality of model $m$, $d_x \sim \mathcal{N}(0, \sigma_d^2)$ is the inherent difficulty of item $x$, and $\varepsilon_{x,m} \sim \mathcal{N}(0, \sigma_\varepsilon^2)$ is independent observation noise.

Our objective is to estimate the quality difference $\Delta_{m,m'} = q_m - q_{m'}$. The unbiased estimator $\hat{\Delta}$ is the difference of the empirical means over the respective evaluation sets $R_m$ and $R_{m'}$:

$$\begin{aligned} \hat{\Delta} &= \hat{\mu}_m - \hat{\mu}_{m'} \\ &= \frac{1}{|R_m|}\sum_{x \in R_m} r_{x,m} - \frac{1}{|R_{m'}|}\sum_{x \in R_{m'}} r_{x,m'} \end{aligned} \tag{30}$$

The variance of this estimator expands to:

$$\mathrm{Var}\left(\hat{\Delta}\right) = \mathrm{Var}(\hat{\mu}_m) + \mathrm{Var}(\hat{\mu}_{m'}) - 2\ \mathrm{Cov}(\hat{\mu}_m, \hat{\mu}_{m'}) \tag{31}$$

Assuming $\varepsilon$ terms are independent across models and items, the covariance term arises solely from the shared item difficulty components $d_x$ present in the intersection $R_m \cap R_{m'}$:

$$
\begin{aligned}
\mathrm{Cov}(\hat{\mu}_m, \hat{\mu}_{m'}) &= \mathrm{Cov}\left(\frac{1}{|R_m|}\sum_{x\in R_m} d_x, \frac{1}{|R_{m'}|}\sum_{x\in R_{m'}} d_x\right) \\
&= \frac{1}{|R_m|\,|R_{m'}|}\sum_{i\in R_m}\sum_{j\in R_{m'}} \mathrm{Cov}(d_i, d_j) \\
&= \frac{1}{|R_m|\,|R_{m'}|}\sum_{x\in R_m\cap R_{m'}} \mathrm{Var}(d_x) \\
&= \frac{|R_m \cap R_{m'}|}{|R_m|\,|R_{m'}|}\sigma_d^2
\end{aligned}
\tag{32}
$$

To strictly minimize $\mathrm{Var}\big(\hat{\Delta}\big)$, we must maximize the covariance term. For any fixed budget sizes $|R_m|$ and $|R_{m'}|$, this term is maximized when the intersection $|R_m \cap R_{m'}|$ is maximized. This condition is met if and only if the evaluation sets satisfy the subset property (i.e., $R_m \subseteq R_{m'}$ or $R_{m'} \subseteq R_m$). ■

**Lemma 3** (Unbiased item ordering): If the item difficulty is strictly monotonic with respect to the item ordering (i.e., items get progressively harder or easier), then the simple average estimator $\hat{\mu}_m$ is a biased estimator of true quality $q_m$ when comparing models with different budget allocations.

**Proof**: Let the score be $r_{x_i,m} = q_m + d_{x_i}$, where $d_{x_i}$ is the difficulty of the $i$-th item. The simple average estimator for model $m$ evaluated on the first $|R_m|$ items is:

$$
\begin{aligned}
\mathbb{E}[\hat{\mu}_m] &= \mathbb{E}\left[\frac{1}{|R_m|}\sum_{k=1}^{|R_m|}\left(q_m + d_{x_i}\right)\right] \\
&= q_m + \frac{1}{|R_m|}\sum_{k=1}^{|R_m|}\mathbb{E}\left[d_{x_i}\right]
\end{aligned}
\tag{33}
$$

Let $\bar{d}_n = \frac{1}{n}\sum_{k=1}^{n}\mathbb{E}\left[d_{x_i}\right]$ be the average expected difficulty of the prefix of length $n$. Now consider two models $m$ and $m'$ with true qualities $q_m = q_{m'} = q$, but unequal budget allocations $|R_m| < |R_m|'$. The expected difference is:

$$
\begin{aligned}
\mathbb{E}[\hat{\mu}_a - \hat{\mu}_b] &= \left(q + \bar{d}_{|R_m|}\right) - \left(q + \bar{d}_{|R_m|'}\right) \\
&= \bar{d}_{|R_m|} - \bar{d}_{|R_m|'}
\end{aligned}
\tag{34}
$$

If the ordering is biased such that difficulty increases with index (monotonic increasing $\mathbb{E}\left[d_{x_i}\right]$), then the prefix average $\bar{d}_n$ is strictly increasing in $n$. Consequently, $\bar{d}_{|R_m|} < \bar{d}_{|R_m|'}$, leading to a non-zero bias term where the model with fewer evaluations ($a$) appears superior solely due to the queue ordering.

$$
\mathbb{E}[\hat{\mu}_m - \hat{\mu}_{m'}] \neq 0 \tag{35}
$$

■

**Theorem 1** (Evaluation allocation is NP hard): Even with oracle access to $\mathcal{M} \times \mathcal{X}$, finding a subset $R^\star$ such that $|R^\star| \leq B$ and the model mean estimates based on $R^\star$ maximize $\tau_\omega(\hat{\mu}, \mu)$ is NP hard for arbitrary $\omega$.

**Proof**: We transform the 0-1 Knapsack problem into an instance of evaluation allocation. We first consider the case where evaluation allocation has constant weight $\omega = 1$ and then extend it to $\omega \in \mathbb{Q}^+$.

Given knapsack capacity $C$ and $N$ items with weights $w_n \in \mathbb{N}$ and profits $p_n \in \mathbb{N}$, we construct an evaluation allocation instance. Partition a set of models $\mathcal{M}$ into $N$ isolated groups. Group $n$ contains one target model $m_n$ and $p_n$ dummy models $m^d_{n,j}$. Order the true capabilities such that $\mu_n > \mu^d_{n,j}$, and ensure strict descending order between groups, $\mu_n > \mu_{n+1}$. To isolate the groups empirically, construct the oracle scores such that all items evaluated by group $n$ yield scores within the interval $I_n = [c_n - \varepsilon, c_n + \varepsilon]$, where $I_1 > I_2 > ... > I_N$. Because the intervals are disjoint, all inter-group empirical pairs are correctly ordered regardless of the allocation. The objective $\tau_\omega$ thus reduces to maximizing correctly ordered intra-group pairs.

For every dummy model $m^d_{n,j}$, fix all item scores to $c_n$. Their empirical mean is invariant: $\hat{\mu}^d_{n,j} = c_n$. Any allocation $|R| > 1$ to a dummy model wastes budget. For each target model $m_n$, define the sequence of item scores such that:

- $\hat{\mu}_n < c_n$ for $|R_n| = 1$
- $\hat{\mu}_n > c_n$ for $|R_n| = 1 + w_n$
- $\hat{\mu}_n < c_n$ for all other $|R| \leq B$

Set the budget to $B = |\mathcal{M}| + C$. Every model must receive at least 1 evaluation, consuming $|\mathcal{M}|$ budget. The remaining budget $C$ can be distributed among the target models. Allocating exactly $w_n$ additional items to $m_n$ raises its empirical mean above its $p_n$ dummy models, increasing the objective in $\tau$ by $p_n$. Any other allocation length for $m_n$ yields an empirical mean below $c_n$ and contributes 0 to the objective. Solving the allocation problem is therefore identical to selecting a subset of Knapsack items to maximize profit $\sum p_n$ subject to $\sum w_n \leq C$. Evaluation allocation with constant $\omega$ is therefore NP-hard.

To extend to arbitrary $\omega$, we simply set the number of items in each group to $z \cdot \frac{p_n}{\omega_n}$ where $z$ is such a problem-level constant so that $\frac{p_n}{\omega_n} \in \mathbb{N}$ for all $n$. A reward for solving a specific group for $m_n$ for the allocation algorithm thus remains $p_n$.

■

**Theorem 2** (Consistency of allocation algorithms): Let $\mathcal{M}$ be a finite set of models and assumptions 1 and 2 hold. Let $\pi$ be an allocation policy such that for any step $t$, the selection probability $P_t(m)$ for every model $m$ satisfies $P_t(m) \geq \zeta$ for some fixed constant $\zeta > 0$. Then, the estimated ranking converges to the true ranking as $B \to \infty$.

**Proof**: We first compute the minimum and maximum for all samplers in the weighted sampling. We make use of the fact that the average is based on observations that are in $[0, 1]$ and so is also bounded:

- $\varepsilon$-Greedy: $1 - \varepsilon \geq \omega_m \geq \frac{\varepsilon}{|\mathcal{M}|}$ for fixed $\varepsilon$
- Boltzmann: $\exp\left(\frac{1}{\tau}\right) \geq \omega_m \geq 1$ for fixed $\tau$
- Rank-based: $1 \geq \omega_m \geq \frac{1}{|\mathcal{M}|}$

The minimum probability of sampling a particular model is at least $\frac{\min_m \omega_m}{|\mathcal{M}| \cdot \max \omega_m}$. Because $\omega_m$ is bounded, the minimum probability is also lower-bounded. We exclude UCB because the only way for UCB probability to tend to zero is for a model to already be pulled infinite amount of times.

Let $|R_{m,B}|$ denote the number of times model $m$ has been evaluated under budget $B$. Since the selection probability for every model is bounded below by $\zeta > 0$, the second Borel-Cantelli lemma guarantees that every model is sampled infinitely often:

$$\lim_{B \to \infty} |R_{m,B}| = \infty, \quad \forall m \in \mathcal{M} \tag{36}$$

Let $\hat{\mu}_{m,B}$ be the simple average estimator for model $m$ after $B$ steps. By the Strong Law of Large Numbers, since $|R_{m,B}| \to \infty$, the estimator converges to the true quality:

$$\hat{\mu}_{m,B} \longrightarrow q_m \tag{37}$$

Let $\delta = \min_{m,m' : q_m \neq q_{m'}} |q_m - q_{m'}|$ be the minimum separation between two distinct true model qualities. For the ranking $\hat{R}_B$ to differ from $R^*$, there must exist at least one pair $m, m'$ such that their estimated order is incorrect. This requires the estimation error for at least one model to exceed $\frac{\delta}{2}$. Since $\hat{\mu}_{m,B}$ converges to $q_m$, there exists a finite budget $B^\dagger$ such that for all $B > B^\dagger$, $|\hat{\mu}_{m,B} - q_m| < \frac{\delta}{2}$ for all $m$, implying $\hat{R}_B = R^*$. ■

We now introduce Lemma 4 which is used for proving Theorem 3.

**Lemma 4** (Asymptotic allocation convergence): Let $\pi$ be a sampling policy where at each step $t$, a model $m$ is selected with probability $P_{t(m)}$ proportional to its estimated importance weight $\hat{\omega}_{m,t}$. Assuming the policy guarantees infinite exploration (i.e., $P_{t(m)} > \varepsilon_t > 0$), then the asymptotic allocation converges to the true importance weights:

$$\lim_{B\to\infty} \frac{|R_{m,B}|}{B} = \frac{\omega_m}{\sum_k \omega_k} \tag{38}$$

**Proof**: Because the policy ensures every model is sampled infinitely often as $B \to \infty$, by the Strong Law of Large Numbers, the empirical mean estimator $\hat{\mu}_{m,t}$ converges to the true mean: $\hat{\mu}_{m,t} \to \mu_m$. Because $\omega_m$ is based on $\hat{\mu}_{m,t}$, similarly $\hat{\omega}_m \to \omega_m$ and also the probability of sampling $P_{t(m)} = \frac{\omega_m}{\sum_k \omega_k} \to = \left(\sum_k \omega_k\right) = P^*(m)$.

The number of samples $|R_{m,B}|$ is the sum of Bernoulli trials with probabilities converging to $P^*(m)$. By the limit theorems for sums of independent random variables, the averaged allocation converges to the sampling probability:

$$\lim_{B\to\infty} \frac{|R_{m,B}|}{B} = P^*(m) = \frac{\omega_m}{\sum_k \omega_k} \tag{39}$$

■

**Theorem 3** (Optimality of weighted sampling for weighted $\tau_\omega$): Assume the estimated model score averages $\hat{\mu}_m$ follows an independent Gaussian distribution $\mathcal{N}\left(\mu_m, \frac{\sigma^2}{|R_m|}\right)$. Sampling according to $\sqrt{\omega_m}$ is the optimal strategy for minimizing the weighted (according to $\omega_m$) ranking uncertainty.

**Proof**: The specific optimality in minimizing variance is known as weighted A-optimality and related to Neyman allocation when constrained by a budget.

Let the true quality of model $m \in \mathcal{M}$ be $\mu_m$. Let $\hat{\mu}_m$ be the estimator for model $m$ after $|R_m|$ samples. Let $\omega_m^2$ be the importance weight of correctly ranking model $m$. We assume the estimated score follows an independent Gaussian distribution:

$$\hat{\mu}_m \sim \mathcal{N}\left(\mu_m, \frac{\sigma^2}{|R_m|}\right) \tag{40}$$

We first define a differentiable loss function that proxies the discrete ranking metric $\tau_\omega$. The probability of incorrectly ranking two models $i$ and $j$ (where $\mu_i > \mu_j$) is the probability that their estimated difference $\hat{\mu}_i - \hat{\mu}_j$ is negative. Since $\hat{\mu}$ are independent Gaussians, the difference $\hat{\mu}_i - \hat{\mu}_j$ is also Gaussian:

$$\hat{\mu}_i - \hat{\mu}_j \sim \mathcal{N}\left(\mu_i - \mu_j, \frac{\sigma^2}{|R_i|} + \frac{\sigma^2}{|R_j|}\right) \tag{41}$$

The probability of ranking error is determined by the tail of this distribution:

$$P\left(\hat{\mu}_i < \hat{\mu}_j\right) = \Phi\left(-\frac{\mu_i - \mu_j}{\sqrt{\frac{\sigma^2}{|R_i|} + \frac{\sigma^2}{|R_j|}}}\right) \tag{42}$$

Minimizing the ranking error is analytically intractable due to the $\Phi$ function and pairwise terms. However, the error probability is strictly monotonic with the variance of the estimators. Therefore, to maximize the precision of the ranking for model $m$ weighted by $\omega_m^2$, we minimize the weighted cumulative variance of the estimators. We define the objective loss function as:

$$\sum_{m\in\mathcal{M}} \omega_m^2 \cdot \mathrm{Var}(\hat{\mu}_m) = \sum_{m\in\mathcal{M}} \frac{\omega_m^2 \sigma^2}{|R_m|} \tag{43}$$

We now show that minimizing Equation 43 under a fixed budget $B$ requires sampling each model $m$ proportional to $\omega_m$. Again, we turn to Lagrange multiplier $\lambda$:

$$\sum_{m \in \mathcal{M}} \frac{\omega_m^2 \sigma^2}{|R_m|} + \lambda \left( \sum_{m \in \mathcal{M}} |R_m| - B \right) \tag{44}$$

We take the partial derivative with respect to the sample count $|R_m|$ and set it to zero to find the stationary points.

$$\frac{\partial \quad \sum_{m \in \mathcal{M}} \frac{\omega_m^2 \sigma^2}{|R_m|} + \lambda \left( \sum_{m \in \mathcal{M}} |R_m| - B \right)}{\partial \ |R_m|} = -\frac{\omega_m^2 \sigma^2}{|R_m|^2} + \lambda = 0 \tag{45}$$

Rearrranging the terms yields:

$$\frac{\omega_m^2 \sigma^2}{|R_m|^2} = \lambda \quad \Rightarrow \quad |R_m| = \frac{\sigma}{\sqrt{\lambda}} \omega_m \tag{46}$$

Sum of the counts also needs to hold to satisfy the budget $B$:

$$\begin{aligned} B &= \sum_{m \in \mathcal{M}} |R_m| \\ &= \sum_{m \in \mathcal{M}} \left( \frac{\sigma}{\sqrt{\lambda}} \right) \omega_m \\ &= \left( \frac{\sigma}{\sqrt{\lambda}} \right) \sum_{m \in \mathcal{M}} \omega_m \end{aligned} \tag{47}$$

$$\Rightarrow \left( \frac{\sigma}{\sqrt{\lambda}} \right) = \frac{B}{\sum_k \sqrt{\omega_k}} \tag{48}$$

We substitute this constant back into the expression for $|R_m|$:

$$\begin{aligned} |R_m|^* &= \left( \frac{B}{\sum_k \omega_k} \right) \omega_m \\ &= B \cdot \frac{\omega_m}{\sum_{k \in \mathcal{M}} \omega_k} \end{aligned} \tag{49}$$

The optimal number of samples $|R_m|^*$ is then proportional to the square root of the importance weight $\omega_m$, which is attained asymptotically by Rank-sqrt weighted sampler according to Lemma 4. ■

**Input**: models $\mathcal{M}$, budget $B$
**Hyperparameter**: warmup $C \in \mathbb{N}$, weight $\omega_m : \mathcal{M} \to \mathbb{R}^+$
**Output**: evaluation results $R$

**WeightedSampling** $(\mathcal{M}, B, C)$:
1 $R \leftarrow \varnothing$
2 **for** $m \in \mathcal{M}$:
3   **repeat** $C$:
4     $x \leftarrow \min_{\prec}\{x \,|\, x \in \mathcal{X}, \langle x, m\rangle \notin R\}$
5     $R \leftarrow R \cup \{\langle x, m\rangle\}$
6 **while** $|R| < B$:
7   $m \leftarrow$ sample $m \in \mathcal{M}$ with prob. $\propto \omega_m$
8   $x \leftarrow \min_{\prec}\{x \,|\, x \in \mathcal{X}, \langle x, m\rangle \notin R\}$
9   $R \leftarrow R \cup \{\langle x, m\rangle\}$
10 **return** $R$

Algorithm 1: Allocation policy with weighted sampling and four weighting functions.

**Input**: models $\mathcal{M}$, budget $B$
**Hyperparameter**: warmup $C \in \mathbb{N}$, weight $\omega_m : \mathcal{M} \to \mathbb{R}^+$
**Output**: evaluation results $R$

**ConfusionMinimization** $(\mathcal{M}, B, C)$:
1 $R \leftarrow \varnothing$
2 **for** $m \in \mathcal{M}$:
3   **repeat** $C$:
4     $x \leftarrow \min_{\prec}\{x \,|\, x \in \mathcal{X}, \langle x, m\rangle \notin R\}$
5     $R \leftarrow R \cup \{\langle x, m\rangle\}$
6 **while** $|R| < B$:
7   $m \leftarrow \text{argmax}_{a \in \mathcal{M}}$ ... ↴
$$\omega_a \sum_{b \in \mathcal{M}} \omega_b \cdot \Phi\left(\frac{|\mu_a - \mu_b|}{\sqrt{\frac{\sigma_a^2}{|R_a|} + \frac{\sigma_b^2}{|R_b|}}}\right) - \omega_b \cdot \Phi\left(\frac{|\mu_a - \mu_b|}{\sqrt{\frac{\sigma_a^2}{|R_a|+1} + \frac{\sigma_b^2}{|R_b|}}}\right)$$
8   $x \leftarrow \min_{\prec}\{x \,|\, x \in \mathcal{X}, \langle x, m\rangle \notin R\}$
9   $R \leftarrow R \cup \{\langle x, m\rangle\}$
10 **return** $R$

Algorithm 2: Allocation policy with confusion minimization.

**Input**: models $\mathcal{M}$, budget $B$
**Output**: evaluation results $R$

**GreedyOracle** $(\mathcal{M}, B)$:
1 $R \leftarrow \varnothing$
2 **while** $|R| < B$:
3   **for** $m \in \mathcal{M}$:
4     $x_m \leftarrow \min_{\prec}\{x \,|\, x \in \mathcal{X}, \langle x, m\rangle \notin R\}$
5   $m^* \leftarrow \text{argmax}_{m \in \mathcal{M}} \tau_\omega(R \cup \{\langle x_m, m\rangle\}, \mathcal{X} \times \mathcal{M})$
6   $R \leftarrow R \cup \{\langle x_{m^*}, m^*\rangle\}$
7 **return** $R$

Algorithm 3: Allocation policy with greedy oracle that has access to full $\mathcal{X} \times \mathcal{M}$ when selecting which model to evaluate next.

**Input**: models $M$, budget $B$
**Output**: evaluation selection $R$

**Uniform** $(M, B)$:
1 $R = \varnothing$
2 **for** $m \in \mathcal{M}$:
3   **repeat** $\frac{B}{|\mathcal{M}|}$:
4     $x \leftarrow \min_{\prec}\{x \,|\, x \in \mathcal{X}, \langle x, m\rangle \notin R\}$
5     $R \leftarrow R \cup \{\langle x, m\rangle\}$
6   **return** $R$

Algorithm 4: Baseline evaluation selection. All models are evaluated on the same set of items.

# C Other Experiments

## C.1 Other Algorithms

In this section we introduce several additional algorithms and in Appendix C.2 discuss their failure in achieving our objectives.

**Statistical Ambiguity**. A good evaluation policy must navigate a trade-off between two distinct sources of ambiguity: variance of a model's score distribution (Cohn et al., 1996), and pairwise indiscriminability which presents itself as low statistical power between adjacent models in the partial ranking (Card et al., 2020). Concretely, this strategy operationalizes the to-be-reduced "ambiguity" as the weighted sum of its confidence interval width and the $p$-values separating it from its neighbors. The reason to consider both at the same time is that minimizing the variance alone is not directly aligned with any of the policy utilities (Section 3), or can lead to overfocusing on a single high-variance model. Similarly, reducing pairwise ambiguity could lead to overfocusing on two models that are naturally very close to each other. We therefore minimize their combined ranks, in order to match their scales.

**Successive Rejects**. Grounded in the framework of best-arm identification, this policy operates in phases, discarding the worst-performing model at each step to concentrate the budget on top-ranking candidates. We appeal to Audibert and Bubeck (2010) for theoretical guarantees on identification of the optimal model. Similar methods exist with different scheduling, such as successive halving (Jamieson and Talwalkar, 2016).

**Successive Halving** is a bracket-based elimination algorithm (Karnin et al., 2013). It operates in rounds, where in each round the budget is equally distributed among the surviving models. After evaluation, the models are ranked by their empirical mean performance, and the bottom half is discarded. This process repeats until a single model remains or the budget is exhausted. Weighted sampling has an advantage over successive rejects or successive halving: when running weighted sampling we can dynamically continue even with apriori unknown budget.

**P-value Rejects** is a dynamic elimination strategy that continuously monitors the pairwise statistical significance between models (Even-Dar et al., 2006). It proceeds in a round-robin fashion, evaluating all active models. After each round, it compares the worst-performing model with the next-worst model using a statistical test (e.g., t-test). If the p-value is below a significance threshold (e.g., 0.05), the worst model is discarded. This allows for early pruning of clearly inferior models without waiting for fixed phases.

**Thompson Sampling** is a Bayesian approach to the exploration-exploitation trade-off (Thompson, 1933). It maintains a posterior distribution for the mean capability of each model. In each step, it samples a potential mean from each model's posterior (approximated as Gaussian) and selects the model (or top-k models) with the highest sampled value for the next evaluation. This naturally balances exploring models with high uncertainty and exploiting models with high estimated performance.

### C.2 Other Objectives

In this section we consider additional objectives to investigate the trade-offs between various allocation algorithms. The simplest objective is the correlation between model ranking given $R_{T_\pi}$ against "true model ranking" given by the full $\mathcal{X} \times \mathcal{M}$. We choose **standard Kendall's $\boldsymbol{\tau}$** variant b, simpler version of the weighted Kendall's $\tau$. An alternative to comparing to final ranking given by all data is comparing to other rankings based on partial annotations given by policy $\pi$. We dub this **stability $\mathtt{Stab.}(\tau_\omega)$** $\stackrel{\text{def}}{=} \mathbb{E}[\tau_\omega(\hat{\mu}, \hat{\mu}')]$, which measures how sensitive the algorithm-predicted model ranking is to stochasticity.

Many evaluations also care about the statistical power of the results and maximizing the number of statistical clusters. We compute the **average $\boldsymbol{p}$-value** between neighboring ranked models using two-sided t-test for related samples (intersection on items that were evaluated by both models). We include the **average payoff** which is very common for the multi-armed bandit. We intentionally consider solely the unbiased item ordering because otherwise the payoff can be hacked by seleting easy items irrespective of the model. Lastly, we include **evaluation focus** $\sum_{m \in \mathcal{M}} \omega \cdot \log|R_m|$, which measures how many evaluations we allocated to the top ranking models, which is useful for collecting reward data.

The results are shown in Table 2. Ambiguity reduction successfully lowers the average p-value but fails to improve ranking correlations. It also presents a confounder: model comparisons with by-chance lower $p$-value are not evaluated which prevents them from raising their $p$-value. For evaluation focus and average payoff, the ambiguity reduction also fails, as it can spend its evaluation budget on any high-variance model regardless of its rank.

### C.3 Other weighting functions $\omega_m$

We also investigate other weighting functions $\omega$ for $\tau$, which correspond to rank position having variying importance for the evaluation objective:

- Top-k $\quad \omega_m = \begin{cases} 1 & \text{if } \mathrm{rank}_m \leq k \\ \frac{1}{|\mathcal{M}|-k} & \text{otherwise} \end{cases}$
- Harmonic1 $\quad \omega_m = \frac{1}{\mathrm{rank}_m^1}$
- Harmonic2 $\quad \omega_m = \frac{1}{\mathrm{rank}_m^2}$
- Harmonic$\frac{1}{2}$ $\quad \omega_m = \frac{1}{\sqrt{\mathrm{rank}_m}}$
- Rev-Harmonic1 $\quad \omega_m = \frac{1}{|\mathcal{M}| - \mathrm{rank}_m^1}$ (worse models preferred)

The results are shown also in Table 2. We include $\sqrt{\omega}$ for each of the variant and also greedy oracle that optimizes directly for this objective. The results show that Sampling can be adapted to be useful under most circumstance, even when the polarity is reversed.

| | $\tau_\omega$ | $\tau$ | **Stability of $\tau_\omega$** | **Eval. focus** | **$p$-val** | $\mu$ | $\tau_\omega \frac{1}{\text{rank}_m^1}$ | $\tau_\omega \frac{1}{\sqrt{\text{rank}_m}}$ | $\tau_\omega$ **top3** | $\tau_\omega \frac{1}{\lvert\mathcal{M}\rvert - \text{rank}_m}$ |
|---|---|---|---|---|---|---|---|---|---|---|
| Uniform | 0.924 | 0.892 | 0.904 | 64.8 | 0.472 | 74.7 | 0.902 | 0.894 | 0.905 | 0.920 |
| Uniform (sampling) | 0.922 | 0.889 | 0.900 | 64.8 | 0.460 | 74.7 | 0.900 | 0.892 | 0.903 | 0.917 |
| Successive rejects | 0.949 | 0.878 | 0.933 | 72.1 | 0.450 | 77.4 | 0.914 | 0.894 | 0.923 | 0.879 |
| Sampling rank | 0.951 | 0.885 | 0.934 | 74.8 | 0.453 | 77.2 | 0.918 | 0.899 | 0.926 | 0.897 |
| Sampling rank$^2$ | 0.948 | 0.853 | 0.922 | 75.1 | 0.456 | 78.3 | 0.903 | 0.876 | 0.915 | 0.850 |
| Sampling $\sqrt[2]{\text{rank}}$ | 0.942 | 0.891 | 0.925 | 72.1 | 0.454 | 76.2 | 0.914 | 0.899 | 0.920 | 0.910 |
| Sampling $\sqrt[4]{\text{rank}}$ | 0.944 | 0.891 | 0.926 | 72.1 | 0.454 | 76.2 | 0.915 | 0.900 | 0.921 | 0.910 |
| Sampling rank $\sqrt{\text{top 3}}$ | 0.945 | 0.887 | 0.926 | 72.4 | 0.453 | 76.1 | 0.913 | 0.897 | 0.921 | 0.908 |
| Sampling rank$^1$ (rev) | 0.874 | 0.872 | 0.831 | 51.0 | 0.456 | 71.8 | 0.862 | 0.864 | 0.861 | 0.922 |
| Sampling $\varepsilon$-greedy | 0.947 | 0.873 | 0.923 | 74.8 | 0.461 | 77.5 | 0.909 | 0.888 | 0.918 | 0.889 |
| Sampling Bolzmann | 0.943 | 0.891 | 0.927 | 71.5 | 0.459 | 77.5 | 0.916 | 0.901 | 0.922 | 0.905 |
| Upper confidence bound | 0.943 | 0.898 | 0.928 | 70.7 | 0.463 | 76.8 | 0.918 | 0.906 | 0.923 | 0.915 |
| Confusion minimization | 0.950 | 0.881 | 0.932 | 72.9 | 0.456 | 77.7 | 0.917 | 0.897 | 0.926 | 0.883 |
| Greedy oracle | 0.951 | 0.894 | 0.934 | 70.2 | 0.483 | 77.1 | 0.923 | 0.906 | 0.930 | 0.903 |
| Greedy oracle rank$^1$ | 0.949 | 0.900 | 0.932 | 68.6 | 0.480 | 76.6 | 0.923 | 0.909 | 0.930 | 0.912 |
| Greedy oracle $\sqrt{\text{rank}}$ | 0.942 | 0.904 | 0.923 | 65.6 | 0.478 | 76.0 | 0.921 | 0.910 | 0.925 | 0.920 |
| Greedy oracle rank top 3 | 0.949 | 0.897 | 0.931 | 68.7 | 0.479 | 76.6 | 0.922 | 0.907 | 0.930 | 0.911 |
| Greedy oracle rank$^1$ (rev) | 0.904 | 0.897 | 0.872 | 54.1 | 0.474 | 73.9 | 0.892 | 0.892 | 0.890 | 0.933 |
| Ambiguity reduction $\lambda$=1 | 0.869 | 0.877 | 0.817 | 51.1 | 0.412 | 74.9 | 0.867 | 0.871 | 0.866 | 0.907 |
| Ambiguity reduction $\lambda$=0 | 0.810 | 0.817 | 0.728 | 45.9 | 0.416 | 75.5 | 0.810 | 0.813 | 0.808 | 0.840 |
| Ambiguity reduction $\lambda$=$\infty$ | 0.872 | 0.869 | 0.826 | 54.2 | 0.449 | 73.9 | 0.863 | 0.863 | 0.862 | 0.910 |
| Thompson sampling | 0.922 | 0.880 | 0.897 | 66.4 | 0.451 | 76.5 | 0.897 | 0.886 | 0.901 | 0.902 |
| $p$-value rejects | 0.931 | 0.898 | 0.912 | 65.7 | 0.466 | 76.0 | 0.910 | 0.902 | 0.913 | 0.922 |
| Successive halving | 0.937 | 0.834 | 0.912 | 73.2 | 0.486 | 78.2 | 0.884 | 0.854 | 0.898 | 0.854 |

Table 2: Selection with unbiased items with different evaluation objectives. Higher is better except for $p$-val.

| | | **Homos.** | **Heteros.** | **Binary** | **Likert** |
|---|---|---|---|---|---|
| **Mean estimator** | Uniform | 0.937 | 0.921 | 0.938 | 0.939 |
| | Uniform (sampling) | 0.933 | 0.921 | 0.935 | 0.929 |
| | Successive rejects | 0.957 | 0.939 | 0.961 | 0.953 |
| | Sampling rank | 0.977 | 0.970 | 0.979 | 0.970 |
| | Sampling $\varepsilon$-greedy | 0.971 | 0.960 | 0.972 | 0.969 |
| | Sampling Bolzmann | 0.936 | 0.916 | 0.935 | 0.927 |
| | Upper confidence bound | 0.935 | 0.921 | 0.939 | 0.942 |
| | Confusion minimization | 0.975 | 0.973 | 0.979 | 0.975 |
| **Linear estimator** | Uniform | 0.932 | 0.924 | 0.938 | 0.939 |
| | Uniform (sampling) | 0.936 | 0.922 | 0.935 | 0.929 |
| | Successive rejects | 0.968 | 0.946 | 0.961 | 0.953 |
| | Sampling rank | 0.983 | 0.969 | 0.979 | 0.970 |
| | Sampling $\varepsilon$-greedy | 0.971 | 0.962 | 0.972 | 0.969 |
| | Sampling Bolzmann | 0.935 | 0.925 | 0.935 | 0.927 |
| | Upper confidence bound | 0.941 | 0.925 | 0.939 | 0.942 |

Table 3: Results ($\tau_\omega$) for synthetic data across four scenarios.

### C.4 Synthetic scale beyond machine translation

In this section we evaluate the algorithms on synthetic datasets, which allow us to control the specific noise profile, as well as scale beyond the limitation of simulations on historical data. For example, we include the common question answering scenario where the output is either correct or incorrect, $r_{x,m} \in \{0, 1\}$ or where the output is scored on Likert-5 scale, $r_{x,m} \in \{\frac{0}{4}, \frac{1}{4}, \frac{2}{4}, \frac{3}{4}, \frac{4}{4}\}$. The data is synthesized with a generative process in four conditions. In all cases we use $|\mathcal{M}| = 50$ models and $|\mathcal{X}| = 500$ items.

- **Homoscedastic:**
  - $\mu_m \sim \mathcal{N}(\mu{=}0.7, \sigma{=}0.25)$
  - $d_x \sim \mathcal{N}(\mu{=}0, \sigma{=}1)$
  - $r_{x,m} \sim \text{clip}(\mu_m + d_x + \mathcal{N}(\mu{=}0, \sigma{=}\bar{\mu}), 0, 1)$ (noise variance constant for all models)
- **Heteroscedastic:**
  - $\mu_m \sim \mathcal{N}(\mu{=}0.7, \sigma{=}0.25)$
  - $d_x \sim \mathcal{N}(\mu{=}0, \sigma{=}1)$
  - $r_{x,m} \sim \text{clip}(\mu_m + d_x + \mathcal{N}(\mu{=}0, \sigma{=}\mu_m), 0, 1)$ (noise variance greater for better models)
- **Binary:**
  - $\mu_m \sim \mathcal{N}(\mu{=}0.7, \sigma{=}0.25)$
  - $d_x \sim \mathcal{N}(\mu{=}0, \sigma{=}1)$
  - $r_{x,m} \sim \text{digitize}(\mu_m + d_x + \mathcal{N}(\mu{=}0, \sigma{=}\mu_m), \langle 0, 1 \rangle)$ (digitize to either 0 or 1)
- **Likert:**
  - $\mu_m \sim \mathcal{N}(\mu{=}0.7, \sigma{=}0.25)$
  - $d_x \sim \mathcal{N}(\mu{=}0, \sigma{=}1)$
  - $r_{x,m} \sim \text{digitize}\big(\mu_m + d_x + \mathcal{N}(\mu{=}0, \sigma{=}\mu_m), \langle \frac{0}{4}, \frac{1}{4}, \frac{2}{4}, \frac{3}{4}, \frac{4}{4} \rangle\big)$ (digitize to five values)

Each scenario is evaluated as average across 100 seeds. The results are shown in Table 3. Across all four scenarios, sampling by the $\omega$ rank is the best choice, also with linear estimation of missing values.

## D Empirical verification of assumptions

The optimality of weighted sampling (Theorem 3) assumes that the mean estimates $\hat{\mu}_m$ are normally distributed and homoscedastic. We empirically evaluate these assumptions on the large-scale WMT data.

Individual evaluation item scores are bounded in $[0, 100]$ and roughly follow a Beta distribution (Figure 9). The sample means $\hat{\mu}_m$ of Beta-distributed items then converge to a Normal distribution. Figure 10 confirms that the empirical distributions of model means across campaigns can be approximated by Normal distributions. Furthermore, the true model capabilities are densely clustered near the upper bound of the scale. This top-heavy density amplifies the statistical ambiguity among leading models, directly supporting the objective of allocating higher budget to top-tier candidates as formulated in Theorem 3.

Theorem 3 also assumes a constant variance $\sigma^2$ across all models. Figure 11 shows slight heteroscedasticity: better-performing models (higher $\mu_m$) have lower variance. This violates the homoscedasticity assumption. We mitigate this imbalance with the Confusion Minimization policy (Algorithm 2). Rather than assuming a uniform $\sigma^2$, this algorithm explicitly incorporates the empirical individual variances ($\sigma_a^2$ and $\sigma_b^2$) into the normal cumulative distribution function to minimize the probability of pairwise ranking errors.

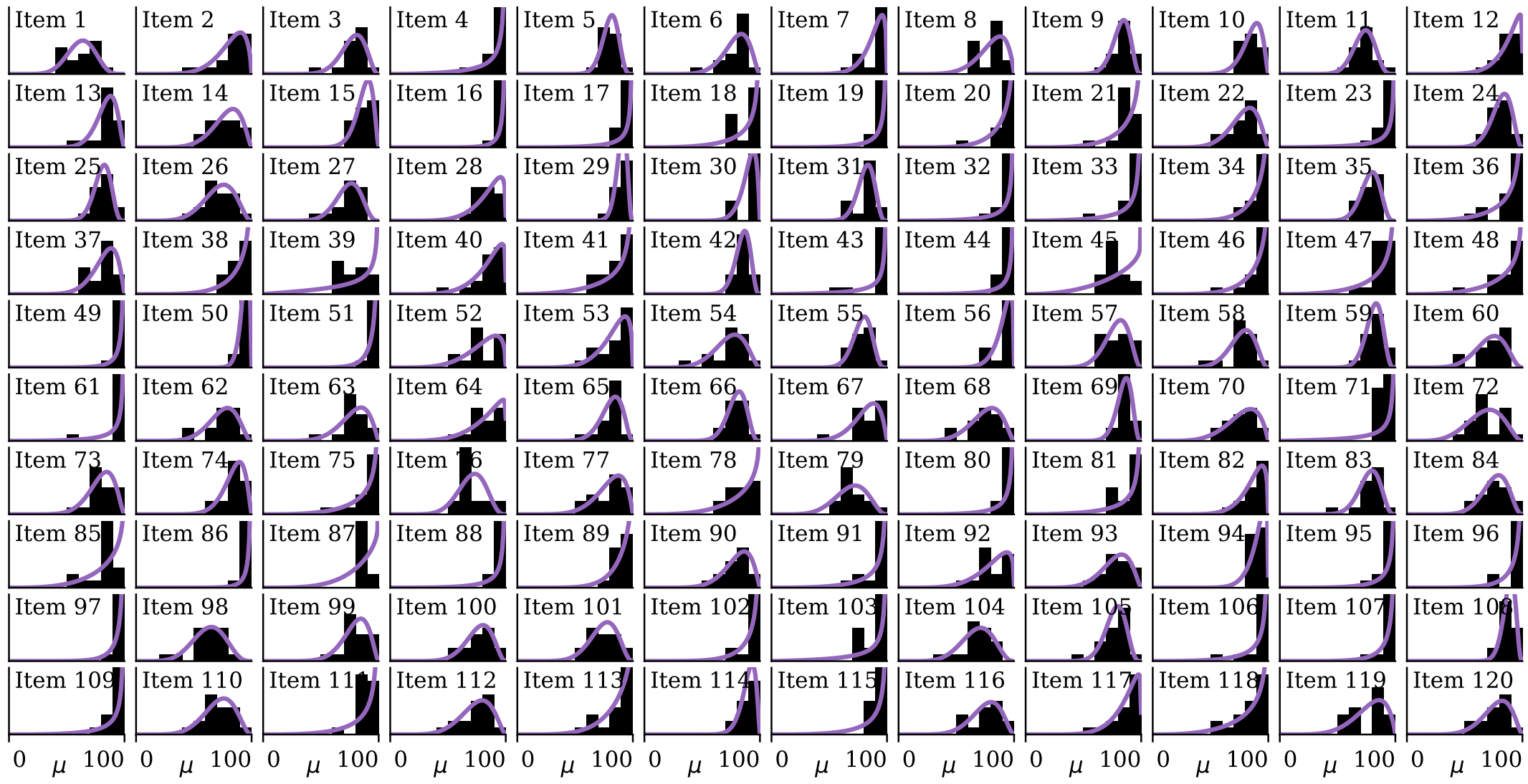

Figure 9: Score distribution for randomly selected items. Overlaying curve shows best matching Beta distribution.

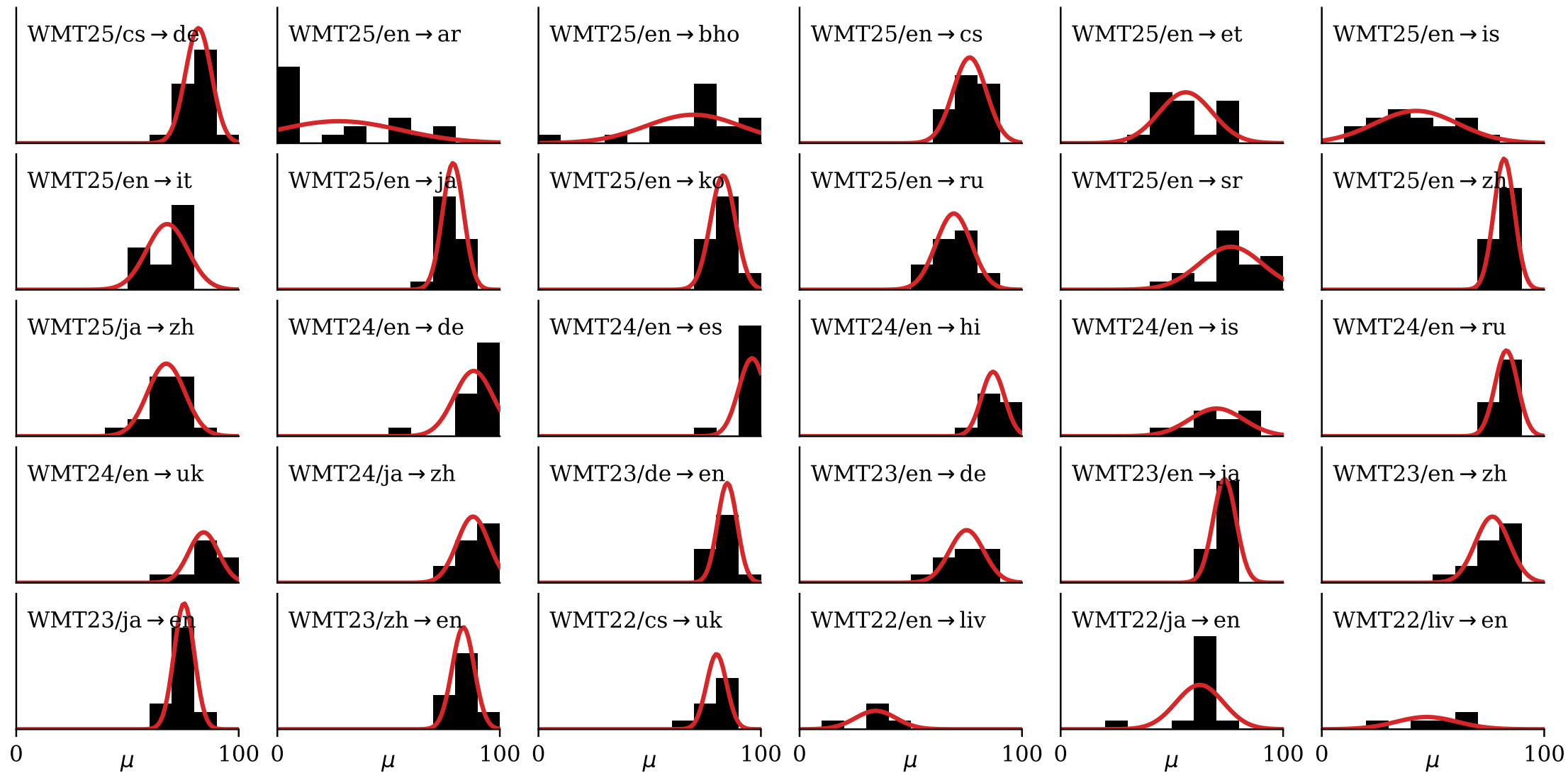

Figure 10: Model mean distribution across datasets. Overlaying curve shows best matching Normal distribution.

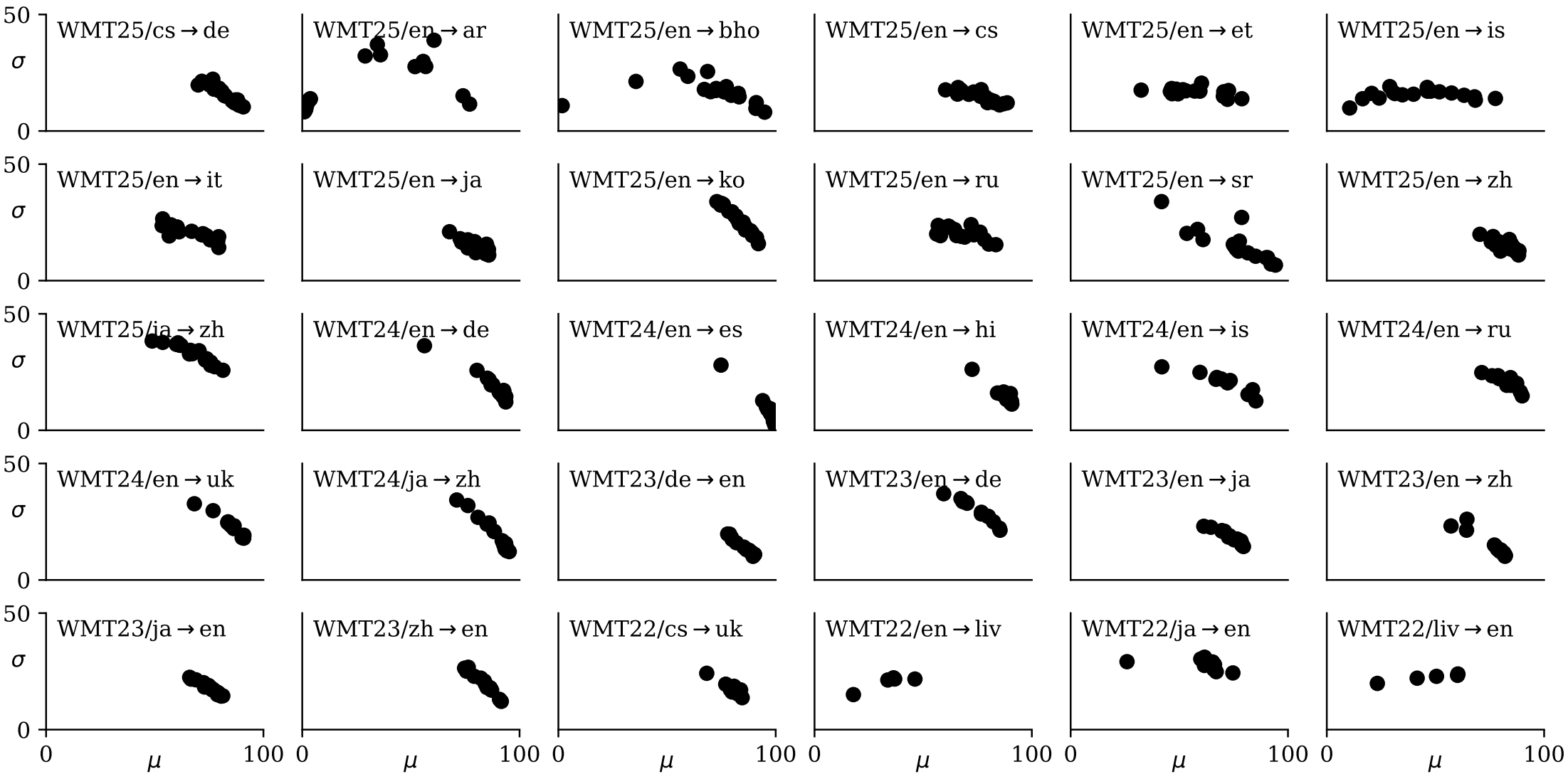

Figure 11: Variance and mean of model scores. The trend shows slight heteroscedasticity: better models tend to have lower variance.

## E Related Works (extended)

This section extends the related work in Section 2 and provides a more in-depth overview of methods related to dynamic evaluation.

**Selection guarantees.** Prior to multi-armed bandits, Bechhofer (1954) defines *indifference zone selection*, which requires that the mean of the selected distribution ("arm") is at most $\delta$ (indifference zone) lower than the true maximum, with a pre-specified probability threshold. This needs to be done with the smallest possible number of samples. Similarly, Gupta (1965) turns this problem into selecting a set of distributions that are guaranteed with a probability threhold to contain the best one, again with the smallest possible number of samples.

These findings were applied to optimal computing budget allocation (Chen et al., 2000) to determine the best distribution. The key finding is that the number of samples is proportial to the variance of the particular distribution and inversely proportional to the gap between the best alternative and this distribution. Distributions with high variance and proximity to the best alternative are thus prioritized.

**Multi-armed bandits.** Modern machine learning has reframed evaluation as a "best arm identification" problem. Audibert and Bubeck (2010); Karnin et al. (2013); Jamieson and Talwalkar (2016) introduce algorithms for fixed-budget best arm identification, such as successive rejects or successive halving. LUCB (Lower Upper Confidence Bound) (Kalyanakrishnan et al., 2012) refines this by sampling the most ambiguous pair: the current best arm and the strongest contender. This algorithm is also formally shown to be good for selecting multiple top models.

Standard algorithms assume that the distribution behind each arm is static and not related to the number of pulls. In our case, however, $i$-th pull for model $m$ is correlated with $i$-th pull for model $m'$. This is known as correlated arms and Gupta et al. (2021) use pseudo-rewards to adapt the above algorithms. Similarly, we use the competition models to estimate the missing scores. In evaluation contexts Schram et al. (2023) use low-rank matrix factorization to infer missing scores.

**Adaptive testing.** Item response theory (IRT, Lord and Novick, 1968) provides a foundation for determining expected student success on a particular exam item by modelling the item difficulty, discriminability, and feasability. Commonly the most discriminative items are selected to be part of a testset (Rodriguez et al., 2021). TrueSkill and ELO (Elo, 1978; Herbrich et al., 2006; Minka et al., 2018), while methodologically different, provide a similar output by modelling the likelihood of winning a match.

Having a model that preditcs success is useful in determining who should have a match with whom and on what evaluation items. This principle guides many recent works (Sakaguchi et al., 2014; Chiang et al., 2024; Balkır et al., 2026) which prioritize matches and evaluation items with the highest uncertainty.

**Application in NLP.** Bouneffouf and Feraud (2025) provide a general overview of bandit algorithms and their applications for large language models, including their evaluation. Moss et al. (2019) use successive halving and a variation of Thompson sampling for best model selection (top-1) from 8 sentiment analysis models. Mohankumar and Khapra (2022); Lanctot et al. (2026) use duelling version of the upper confidence bound bandit algorithm for annotation tasks of pairwise comparison, including translation. Shi et al. (2024) apply multi-armed bandit for efficient prompt learning. Zhang and Duh (2024) make use of successive halving for resource allocation during translation model training. Different from bandit approach, Son et al. (2025); Yuksel et al. (2026) rely on multiple elimination tournaments for model selection. The focus of Zhou et al. (2025) is top-1 selection, which would correspond to specific $\omega_1 = 0$ and $\omega_{>1} = 0$ in our case. Lastly, Li et al. (2025a); Wang et al. (2025); Li and Xiong (2025); Wang et al. (2026) revisit evaluation item selection for LLMs based on distributional properties. Concurrently, Saha et al. (2026) examine optimal allocation for repeated LLM-based evaluation with high item-wise variance and noise.